\documentclass[runningheads]{llncs}

\usepackage[mobile]{eccv}

\usepackage{eccvabbrv}

\usepackage{graphicx}
\usepackage{booktabs}
\usepackage[normalem]{ulem}
\usepackage[accsupp]{axessibility}

\usepackage{algorithm}
\usepackage{algpseudocode}
\usepackage{kotex}
\usepackage{wrapfig}
\usepackage{graphicx}
\usepackage{subcaption}

\usepackage[hidelinks]{hyperref}

\usepackage{orcidlink}

\begin{document}

\title{When Sinks Help or Hurt: Unified Framework for Attention Sink in Large Vision-Language Models}
\titlerunning{Unified Framework for Attention Sink in LVLMs}

\author{Jiho Choi\orcidlink{0000-0002-7793-4122}\inst{1} \and Jaemin Kim\orcidlink{0009-0008-4463-3586}\inst{2}\\[2pt]
Sanghwan Kim\orcidlink{0009-0008-0919-7256}\thanks{Corresponding authors.}\inst{3} \and
Seunghoon Hong\orcidlink{0009-0004-7685-5352}$^\star$\inst{1} \and
Jin-Hwi Park\orcidlink{0000-0001-7874-2344}$^\star$\inst{2}
}

\authorrunning{J.~Choi et al.}

\institute{KAIST, South Korea \and
Chung-Ang University, South Korea \and
Technical University of Munich, Germany\\
\email{\{jiho.choi,seunghoon.hong\}@kaist.ac.kr}\\
\email{\{kimjm1213,jinhwipark\}@cau.ac.kr}\\
\email{sanghwan.kim@tum.de}}

\maketitle
\setcounter{footnote}{0}% ===== Changed (R3): reset footnote counter so body footnotes start at 1 (title-page \thanks left it at 3) =====

\begin{figure}[t]
    \centering
    \vspace{-10pt}
    \includegraphics[width=\linewidth]{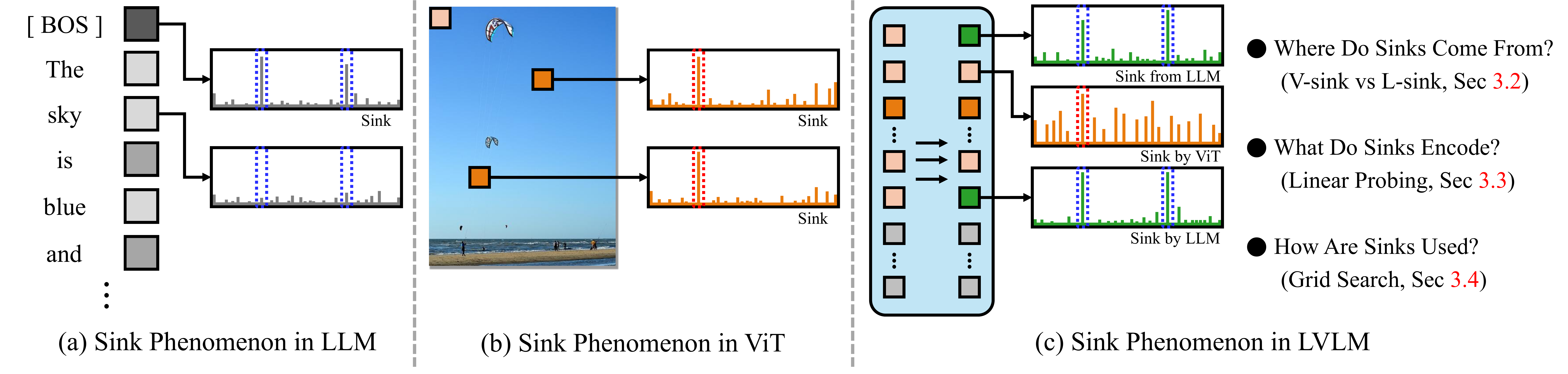}
    \caption{Attention sink phenomena across model modalities.
(a) In LLMs, certain tokens consistently receive disproportionately high attention scores across layers. (b) In ViTs, a similar pattern appears where background patches accumulate high attention. (c) In LVLMs, visual sink tokens among the vision token sequence arise from two distinct sources: those inherited from the vision encoder and those newly formed within the LLM layers, motivating three questions investigated in this work: (1) where sinks originate, (2) what information sinks encode, and (3) how sinks affect downstream task performance.}
    \label{fig:overall} % [2026-06-23 uncommented: kept active for potential cross-reference; teaser currently uncited by design]
\vspace{-15pt}
\end{figure}

\begin{abstract}
Attention sinks are defined as tokens that attract  disproportionate attention. While these have been studied in single modality transformers, their cross-modal impact in Large Vision-Language Models (LVLM) remains largely unexplored: are they redundant artifacts or essential global priors? This paper first categorizes visual sinks into two distinct categories: ViT-emerged sinks (V-sinks), which propagate from the vision encoder, and LLM-emerged sinks (L-sinks), which arise within deep LLM layers. Based on the new definition, our analysis reveals a fundamental performance trade-off: while sinks effectively encode global scene-level priors, their dominance can suppress the fine-grained visual evidence required for local perception. Furthermore, we identify specific functional layers where modulating these sinks most significantly impacts downstream performance. To leverage these insights, we propose Layer-wise Sink Gating (LSG), a lightweight, plug-and-play module that dynamically scales the attention contributions of V-sink and the rest visual tokens. LSG is trained via standard next-token prediction, requiring no task-specific supervision while keeping the LVLM backbone frozen. In most layers, LSG yields improvements on representative multimodal benchmarks, effectively balancing global reasoning and precise local evidence.\footnote{Code will be released at \url{https://github.com/JH-GEECS/lsg_public}}% ===== Changed (R3): code repo footnote =====

\end{abstract}

\section{Introduction}

Transformers are widely known to exhibit the \emph{attention sink} phenomenon, where a small subset of tokens attracts disproportionately large attention weights despite possessing limited semantic content~\cite{xiao2023efficient, guattention, yu2024unveiling, darcetvision}. In Large Language Models (LLMs), these sinks often manifest in punctuation or special symbols (e.g., \texttt{[BOS]} token) and are often characterized by massive activations in specific hidden dimensions~\cite{sun2024massive, yu2024unveiling}. A similar behavior exists in Vision Transformers (ViTs), where background patches with high $\ell_2$ norms absorb attention to facilitate global context propagation~\cite{darcetvision, jiangvision, lappe2025register}. While recent works~\cite{wang2025mirage, srikrishnan2025blindsight, kang2025see, luo2025sink, kanrar, bai2025self} have begun to investigate how these tokens interact within multimodal architectures (e.g., LLaVA~\cite{liu2023visual}), it remains unexplored whether they represent undesirable artifacts to be removed or essential components encoding useful global priors.

To clarify the role of sink tokens in multimodal tasks, we first categorize them into two fundamentally distinct types based on their computational origins: \emph{ViT-emerged sinks} (V-sinks), which originate from the vision encoder and persist as high-norm representations, and \emph{LLM-emerged sinks} (L-sinks), which arise within the deeper layers of the LLM through sharp, dimension-specific activations. While both types consistently attract significant attention during generation, they stem from different computational origins and display unique activation patterns. 
Our analysis of the information encoded within V-sinks, L-sinks, and ordinary tokens reveals a distinct distribution of semantic attributes across each category such as count, size, color, and shape. This suggests that these token types can be actively utilized depending on the character of downstream task. To validate this, we perform oracle interventions by modulating the Key scaling of specific token groups. These experiments expose a fundamental trade-off: while sinks are vital for tasks reliant on global context, they can become detrimental to local tasks requiring fine-grained perception. Furthermore, we identify specific \emph{functional layers} where modulating these sinks is impactful, suggesting that optimal sink utilization is highly sensitive to both task requirements and layer depth.

To leverage these insights, we propose \textbf{Layer-wise Sink Gating (LSG)}, a lightweight, plug-and-play module that dynamically adjusts the attention contributions of V-sinks, L-sinks, and ordinary tokens. Conditioned on the hidden state of the final input token, which aggregates both visual context and query semantics~\cite{kim2025interpreting,neotowards, zhao2024first}, LSG predicts layer-specific key scaling factors. This small gating module is trained via standard next-token prediction loss while the LVLM backbone remains frozen, requiring no task-specific supervision. Our framework enables the model to balance global knowledge and local perception in a content-aware manner. We evaluate LSG on representative multimodal benchmarks, demonstrating improvements, particularly in vision-centric tasks.

Our contribution can be summarized as follows. (1) We provide a unified analysis of attention sinks in LVLMs, characterizing two distinct categories: \emph{ViT-emerged sinks} (V-sinks) and \emph{LLM-emerged sinks} (L-sinks). (2) We reveal a fundamental trade-off in multimodal tasks: while sink tokens facilitate global scene understanding, ordinary tokens are more effective for fine-grained perception tasks relying on precise local evidence. (3) We introduce \textbf{Layer-wise Sink Gating (LSG)}, a lightweight module that dynamically scales sink contributions, yielding performance gains on representative multimodal benchmarks.

\section{Related Work}
\label{sec:related}

\subsection{Attention Sink in Large Language Models}

The attention sink phenomenon has been observed in transformer-based language models, where disproportionately large attention scores are allocated to a small subset of tokens (e.g., \texttt{[BOS]} token or punctuation tokens) despite their limited semantic significance~\cite{xiao2023efficient, yu2024unveiling, cancedda2402spectral}. Recent work~\cite{sun2024massive} shows that this property arises from massive activations of specific dimensions in the hidden states. Additionally, an investigation of the different characteristics of attention sinks is conducted by~\cite{guattention,ruscio2025you}, which demonstrates that they behave more like key biases rather than contributing to value computation.
Recent works~\cite{han2025zerotuning, su2025kvsink, liusinktrack} suggest that sink tokens serve as indispensable structural biases, motivating the performance improvements without additional training.

\subsection{Attention Sink in Vision Encoders}

For ViT-based vision encoders~\cite{touvron2022deit, oquab2024dinov2, cherti2023reproducible}, a similar phenomenon has also been observed, where high-norm tokens tend to emerge in low-informative background regions. 
Some works~\cite{darcetvision,jiangvision} find that they contain global information across the entire image and propose register tokens to reduce artifacts in the feature maps, which improves dense prediction performance (e.g., semantic segmentation and object detection). In addition, Wang \etal~\cite{lappe2025register} reveals that the \texttt{[CLS]} token tends to rely on these sink tokens, and Lu \etal~\cite{lu2025artifacts} modifies the attention dynamics to take advantage of the different roles of massive and artifact tokens.

\subsection{Attention Sink in LVLMs}

Recent studies~\cite{kang2025see, luo2025sink} have explored the attention sink phenomenon in Large Vision-Language Models (LVLMs) with conflicting interpretations. Kang \etal~\cite{kang2025see} argue that sink tokens are largely irrelevant to performance and propose redistributing their attention via threshold-based head selection. Conversely, Luo \etal~\cite{luo2025sink} suggest these tokens encode vital global information, advocating the active adjustment of sink contributions based on task-specific (global vs. local) requirements. From a reliability perspective, Wang \etal~\cite{wang2025mirage} investigate how sinks may exacerbate hallucination errors and their potential utility in hallucination-based attacks. Given these divergent perspectives, it remains an open question whether attention sinks constitute redundant artifacts to be eliminated or functional anchors to be strategically utilized. To bridge this gap, we provide a unified analysis of attention sinks across both vision and language modalities. We then propose a layer-wise sink-gating mechanism for task-aware modulation. This approach is motivated by recent findings~\cite{shi2025vision} indicating that LLM layers exhibit distinct functional specializations, such as counting, grounding, and OCR, requiring different levels of sink-based information.
A structured comparison with Kang~\etal~\cite{kang2025see} and Luo~\etal~\cite{luo2025sink} is provided in Appendix~\ref{sec:supp_novelty}.

\section{Analysis of Visual Sink Tokens}

\subsection{Preliminaries}
Recent Large Vision-Language Model (LVLM)~\cite{liu2023visual, li2025llavaonevision, Qwen2.5-VL} typically consist of (i) a \textbf{vision encoder} $\mathcal{E}$~\cite{radford2021learning, zhai2023sigmoid}, (ii) a MLP-based \textbf{projector} $\mathcal{P}$ that maps vision tokens to the language embedding space~\cite{merullolinearly, liu2023visual}, and (iii) a \textbf{Large Language Model (LLM)} $\mathcal{M}$~\cite{touvron2023llama, qwen2}.
Given an input image $\mathbf{I}$, the vision encoder outputs $n$ patch-level representations of dimension $D_v$,
$\mathbf{V}=[\mathbf{v}_1,\dots,\mathbf{v}_n] \in \mathbb{R}^{n\times D_v}$.
The projector then maps these representations to match the LLM hidden dimension $D$, producing \textbf{visual tokens}
$\mathbf{X}_{\mathrm{vis}} \in \mathbb{R}^{n\times D}$.
The LLM takes as input a single concatenated sequence of \textbf{system tokens} $\mathbf{X}_{\mathrm{sys}}$, \textbf{visual tokens}
$\mathbf{X}_{\mathrm{vis}}$, and \textbf{query tokens} $\mathbf{X}_{\mathrm{txt}}$, and operates in an autoregressive manner.
In this paper, we analyze how visual tokens—especially those fed into the LLM after the projector—are grouped and utilized
across layers within this LVLM architecture.

\noindent\textbf{Layer-wise Hidden Representations.}\quad
Let the hidden states of the entire input sequence at the $\ell$-th LLM layer be denoted by
$\mathbf{H}^{\ell}\in\mathbb{R}^{T\times D}$, where
$T = |\mathcal{I}_{\mathrm{sys}}| + |\mathcal{I}_{\mathrm{vis}}| + |\mathcal{I}_{\mathrm{txt}}|$
is the total sequence length.
The $\ell$-th layer hidden state can be written with a residual formulation as:
\begin{equation}
    \mathbf{H}^{\ell} \;=\; \mathbf{H}^{\ell-1} + F^{\ell}\!\left(\mathbf{H}^{\ell-1}\right),
    \label{eq:residual}
\end{equation}
where $F^\ell$ is a transformer block that includes self-attention and an MLP (FFN).
All of our layer-wise analyses begin from $\mathbf{H}^{\ell}$, and we introduce additional notation only as needed
based on this definition.

\noindent\textbf{Sink Identification and Token Grouping.}\quad
Recent work~\cite{sun2024massive} observed that Transformer-based models can exhibit abnormally large activation values in specific hidden dimensions. Concretely, such behavior appears consistently in a fixed set of
dimensions within each model across different modality (e.g., dimension 650 in CLIP-L, or dimensions 1415 and 2533 in LLaMA2-7B).
Following~\cite{kang2025see}, we call these dimensions the \textbf{sink dimensions} $\mathcal{D}_{\mathrm{sink}}$.
Building on this finding, we identify \textbf{sink tokens} in both the ViT and the LLM by applying thresholding to the activation
magnitude on sink dimensions (threshold selection and its justification are detailed in Appendix~\ref{sec:supp_sink_id}).

Let $\mathbf{h}_j^{\ell} \in \mathbb{R}^D$ be the hidden state of token $j$ at layer $\ell$.
Given a threshold $\tau$, we define the set of sink-token indices at layer $\ell$ as:
\begin{equation}
    \hat{\mathcal{I}}^{\ell}
    =\left\{\, j\in\mathcal{I}\;\middle|\; \max_{d\in \mathcal{D}_{\mathrm{sink}}}\left|(\mathbf{h}_j^{\ell})[d]\right| \ge \tau \right\}.
    \label{eq:sink_def}
\end{equation}

Referring to this criterion~\cite{kang2025see, luo2025sink}, we partition $\mathcal{I}_{\mathrm{vis}}$ into three groups:

\begin{itemize}
    \item \textbf{ViT-emerged sinks}~\cite{luo2025sink} (V-sink, $\hat{\mathcal{I}}_{\mathrm{vit}}$):
          Token indices that are already classified as sinks in the vision encoder (right before the projector), using a criterion analogous to
          Eq.~(\ref{eq:sink_def}). After being passed into the LLM, we treat them as a fixed set of indices.

    \item \textbf{LLM-emerged sinks}~\cite{kang2025see} (L-sink, $\hat{\mathcal{I}}_{\mathrm{llm}}^{\ell}$):
          Vision tokens that are ordinary at the ViT output stage, but become sink tokens inside the LLM after passing through attention/MLP
          computations, i.e., they satisfy Eq.~(\ref{eq:sink_def}) at some LLM layer $\ell$. These are identified in a layer-wise manner.

    \item \textbf{Ordinary visual tokens} ($\mathcal{I}_{\mathrm{ord}}^{\ell}$):
          The remaining visual tokens that do not belong to the two sink groups, defined as
          $\mathcal{I}_{\mathrm{ord}}^{\ell} = \mathcal{I}_{\mathrm{vis}} \setminus (\hat{\mathcal{I}}_{\mathrm{vit}} \cup \hat{\mathcal{I}}_{\mathrm{llm}}^{\ell})$.
\end{itemize}

\noindent Token group statistics, including average sink counts per sample, are reported in Appendix~\ref{sec:supp_token_stats}.

\subsection{Two Sources of Visual Sinks: ViT vs.\ LLM}
\label{sec: hidden state stat}
\begin{wrapfigure}{r}{0.55\textwidth}
    \vspace{-15pt}
    \centering
    \includegraphics[width=\linewidth]{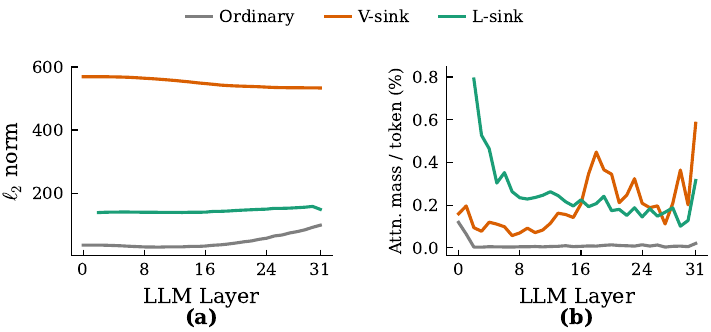}
    \vspace{-20pt}
    \caption{\textbf{Layer-wise salience pattern of visual token groups.} (a) \vsinks{} and \lsinks{} maintain higher $\ell_2$ norms. (b) They also receive a significantly larger percentage of attention mass compared to ordinary visual tokens.}
    \label{fig:salience_norm_attn}
    \vspace{-15pt}
\end{wrapfigure}

Unless otherwise stated, all analyses in this section are conducted with LLaVA-1.5-7B~\cite{liu2023visual} (CLIP ViT-L/14 + Llama 2-7B). The layer-wise norm and attention patterns (Figure~\ref{fig:salience_norm_attn}) and the group-wise activation profiles (Figure~\ref{fig:activation_3d}) are replicated on LLaVA-OneVision-7B (SigLIP + Qwen2) in Appendix~\ref{sec:supp_cross}, confirming that the sink taxonomy generalizes across architectures.

Prior interpretability studies~\cite{neotowards, kaduri2025s, kim2025interpreting} have shown that the final input token's attention over visual tokens critically mediates image-to-text information flow in LVLMs.
We analyze this attention for the three token groups ($\hat{\mathcal{I}}_{\mathrm{vit}}$, $\hat{\mathcal{I}}_{\mathrm{llm}}^{\ell}$, $\mathcal{I}_{\mathrm{ord}}^{\ell}$) using 300 GQA~\cite{hudson2019gqa} samples, measuring the attention weight from the final input token to each group per layer (with head-wise and sample-wise averaging). As shown in Figure~\ref{fig:salience_norm_attn}, both sink groups receive significantly larger per-token attention weights than ordinary tokens, consistent with their abnormally large norms~\cite{darcetvision, sun2024massive}.

Although both sink groups attract strong attention from the final input token, their origins differ.
Prior work has shown that attention sinks arise from massive activations in a few fixed hidden dimensions, leading to large key norms and disproportionate attention weight~\cite{sun2024massive}. The same phenomenon has been observed in both vision transformers~\cite{darcetvision} and LVLMs~\cite{kang2025see}.

\begin{figure}[t]
    \centering
    \vspace{-15pt}
    \includegraphics[width=\linewidth]{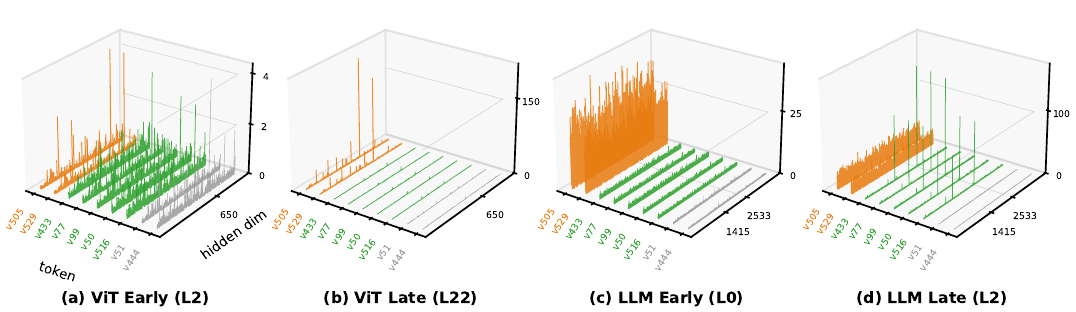}
    \vspace{-15pt}
    \caption{
        \textbf{Activation pattern of visual tokens across LVLM stack} at different depth level. Each line represents one token, colored by category: \textcolor{orange}{\textbf{orange}}~=~ViT-emerged sinks, \textcolor[rgb]{0.17,0.63,0.17}{\textbf{green}}~=~LLM-emerged sinks, \textcolor{gray}{gray}~=~ordinary tokens. Sink dimensions: 650 for CLIP-ViT-L; 1415, 2533 for LLaMA-2-7B. The input image is shown in Figure~\ref{fig:image_overlay}.
    }
    \label{fig:activation_3d}
    \vspace{-15pt}

\end{figure}

V-sinks carry large activation values from the vision encoder and enter the LLM with uniformly high norms (Figure~\ref{fig:activation_3d}~(c))~\cite{luo2025sink}. In the ViT, these tokens develop sharp spikes in a few sink dimensions under bidirectional attention (Figure~\ref{fig:activation_3d}~(a,b)). However, the two-layer MLP projector, which lacks a residual connection, applies a linear transformation followed by a nonlinearity, spreading the concentrated activation across output dimensions. As a result, V-sinks arrive in the LLM embedding space with uniformly elevated norms rather than dimension-specific spikes.
L-sinks, by contrast, appear ordinary at the ViT output but develop sharp dimension-specific activations at deeper LLM layers (Figures~\ref{fig:activation_3d}~(b) and \ref{fig:activation_3d}~(d)). These activations occur in the same $\mathcal{D}_{\mathrm{sink}}$ as text sink tokens (e.g., BOS) and are written by early-layer FFNs and maintained through residual connections, mirroring the text-sink formation mechanism~\cite{kang2025see}.
Because V-sinks and L-sinks follow fundamentally different formation mechanisms, one shaped by ViT bidirectional attention and the projector, the other by the LLM's own text-sink mechanism, they are expected to carry different information despite both attracting strong attention.

\subsection{What Do Visual Sinks Encode? A Linear Probing Approach}

As discussed in Section~\ref{sec:related}, high-norm outlier tokens in vision transformers aggregate global image information~\cite{darcetvision}, and Luo \etal~\cite{luo2025sink} further confirmed this by isolating visual tokens through modified attention masks and mapping them to output vocabulary.
However, both studies characterize sink tokens only at the ViT output or through indirect decoding, without examining how their information content changes across LLM layers. Meanwhile, Kang \etal~\cite{kang2025see} identified \lsinks{} that emerge within the LLM decoder and showed that masking them causes little performance degradation, yet what these \lsinks{} actually encode remains uncharacterized.

Fu \etal~\cite{fuhidden} show that vision representations are broadly preserved across LLM layers, while Feucht \etal~\cite{feucht2024token} demonstrate that token identities can be rapidly erased in early layers. Since visual tokens are projected into the language embedding space via an MLP connector~\cite{merullolinearly}, \vsinks{} may undergo similar restructuring, yet no prior work has examined them separately.

To address this gap, we design a linear probing study on the CLEVR dataset that tracks \vsinks{} ($\hat{\mathcal{I}}_{\mathrm{vit}}$), \lsinks{} ($\hat{\mathcal{I}}_{\mathrm{llm}}^{\ell}$), and ordinary visual tokens ($\mathcal{I}_{\mathrm{ord}}^{\ell}$) across LLM layers, probing four scene-level properties: object \emph{count}, \emph{color}, \emph{shape}, and \emph{size}. These cover different abstraction levels, from holistic scene understanding (counting) to per-object categorical attributes (color, shape) and spatial reasoning (relative size), letting us ask: (1)~what granularity of scene information do sink tokens retain, and (2)~does this information degrade, persist, or strengthen across successive LLM layers?

\noindent\textbf{Experimental Setup.}\quad
We define four tasks spanning different prediction types. Count and Size are multi-class classification problems: Count predicts the total number of objects in the scene, while Size filters the largest single object by bounding-box ratio and assigns it to one of 5 bins. Color and Shape are multi-label classification problems: Color predicts $y\in\{0,1\}^8$ indicating the presence of 8 colors, and Shape predicts $y\in\{0,1\}^3$ indicating the presence of 3 shapes. As a baseline, we construct an ordinary sampled group by excluding all tokens identified as ViT sinks or LLM sinks at any layer, then randomly sampling 5 tokens per image (${\sim}$1\% of 576 patches in CLIP ViT-L/14 336px~\cite{radford2021learning}) from this pool and averaging their hidden states. The same token indices are used across all LLM layers to ensure that layer-wise trends reflect the evolution of identical tokens.
To control for sampling variance, we repeat this construction with five independent random samplings and report the mean with $\pm1$ standard deviation (shaded) for the ordinary group.

\begin{figure}[t]
    \centering
    \vspace{-10pt}
    \includegraphics[width=\linewidth]{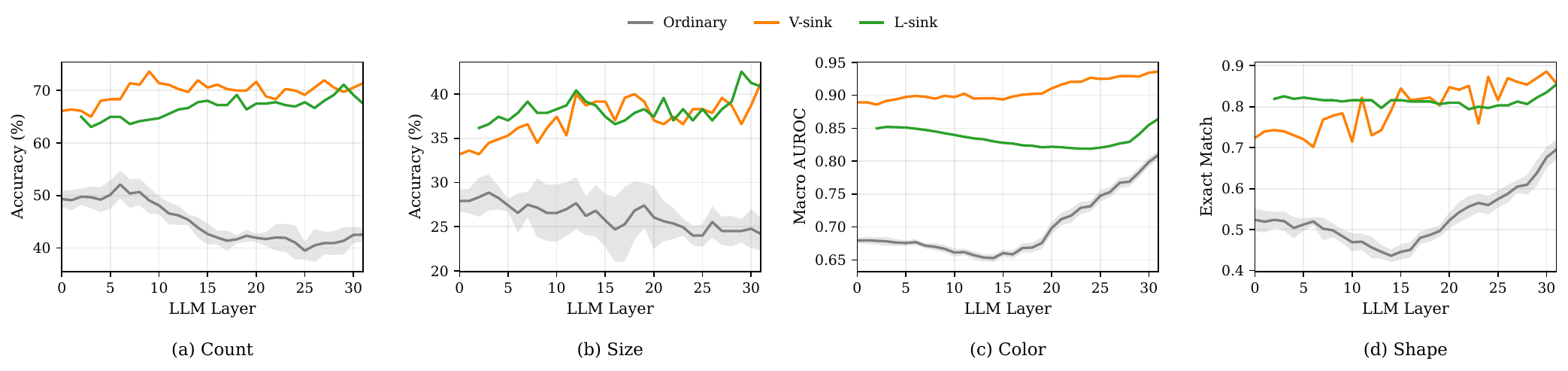}
    
    \vspace{-10pt}
    
    \caption{\textbf{Layer-wise linear probing on CLEVR scene attributes.} Each panel probes one property (count, size, color, shape) from pooled hidden states across LLM layers of LLaVA-1.5-7B: \vsinks{} and \lsinks{} follow the sink criterion of Eq.~\eqref{eq:sink_def} and the V-/L-sink partition defined below it, while the ordinary curve averages five independent random samplings of the ordinary group with $\pm1\sigma$ shading.}
    
    \label{fig:clevr_probing_results}
    \phantomsubcaption\label{fig:clevr_count}%
    \phantomsubcaption\label{fig:clevr_size}%
    \phantomsubcaption\label{fig:clevr_color}%
    \phantomsubcaption\label{fig:clevr_shape}

    \vspace{-20pt}

\end{figure}

\noindent\textbf{Results.}\quad
Figure~\ref{fig:clevr_probing_results} summarizes the layer-wise probing results.
For question (1), both V-sink and L-sink groups outperform ordinary visual tokens across all four tasks, with the advantage most evident in early-to-middle layers. The gap is largest for count prediction (Figure~\ref{fig:clevr_count}), a holistic property that no single patch can capture alone, confirming that sink tokens concentrate scene-level summary information. The advantage also holds for size bin classification (Figure~\ref{fig:clevr_size}), which requires spatial reasoning about relative object scale. For color and shape (Figures~\ref{fig:clevr_color},~\ref{fig:clevr_shape}), sink tokens lead in the early and middle layers, while ordinary tokens gradually close the gap at deeper layers as the LLM's own processing enriches their representations.

For question (2), probing accuracy for sink tokens either remains stable or increases at deeper layers across all four tasks, showing almost no degradation despite the layer-wise transformations. Unlike the token-identity erasure reported for language tokens in early layers~\cite{feucht2024token}, the scene-level information in sink tokens is preserved throughout all LLM layers.
% We further repeat the probing with five random samplings of the ordinary group and observe the same trends across all seeds.

Taken together, these results establish that both V-sinks and L-sinks carry rich, multi-granularity scene information that persists throughout the LLM. This confirms that L-sinks, despite following the text-sink formation mechanism, do not lose the visual information present in their representations. However, the presence of information does not imply that the model actively exploits it during inference~\cite{fuhidden}. We therefore turn to attention-level interventions to examine whether and how the information in each sink type is utilized.

\subsection{Layer-wise Intervention on Sink vs.\ Ordinary Attention}
\label{sec:grid search}

Our probing results show that sink tokens stably encode global attributes across LLM layers. Combined with evidence that visual processing becomes functionally specialized at deeper layers~\cite{shi2025vision}, a natural question arises: {at which layers, and how, does the global information in sinks actually contribute to visual reasoning?}

Prior work probes information flow via hard attention knockouts~\cite{kaduri2025s, geva2023dissecting} or continuous rescaling~\cite{wang2025v}.
The former masks attention connections at selected layers to identify which token-to-token interactions are essential for a correct prediction, whereas the latter multiplies attention weights by learned or fixed scalars and observes the resulting performance change.
Since our question concerns the balance between sink and ordinary attention rather than which specific interactions are essential, we follow the continuous rescaling approach.

Our formulation is closest to Wang \etal~\cite{wang2025v}, but differs in the axis of intervention: Wang \etal~\cite{wang2025v} assigns a scalar to each attention head independently to identify which heads are influential for specific visual semantics, whereas we assign a scalar to each token group and apply it uniformly across all heads within a layer, in order to probe how the relative attention allocation between sink and ordinary tokens affects downstream reasoning at each layer.

\noindent\textbf{Formulation (Key Gating).}\quad
To intervene, we introduce a Key Gating that smoothly scales the attention contribution of a specific token group.
Consider a standard attention operation at layer $\ell$, where the previous hidden state $\mathbf{H}^{(\ell-1)} \in \mathbb{R}^{T \times D}$ is linearly projected into query $\mathbf{Q}^{(\ell)}$, key $\mathbf{K}^{(\ell)}$, and value $\mathbf{V}^{(\ell)}$. For simplicity, we present the single-head form, but the intervention is applied identically to all heads in the layer. To control token group-wise attention contribution, we intervene by multiplying the key vectors by scalar coefficients. Let the token-wise coefficient vector be $\mathbf{s}\in\mathbb{R}^{T}$. The intervened attention is defined as:
\begin{equation}
    \mathrm{Attn}\big(\mathbf{Q}^{(\ell)}, \mathrm{diag}(\mathbf{s}) \cdot \mathbf{K}^{(\ell)}, \mathbf{V}^{(\ell)}\big)
    = \mathrm{SoftMax}\!\left(\frac{\mathbf{Q}^{(\ell)}\big(\mathrm{diag}(\mathbf{s}) \cdot \mathbf{K}^{(\ell)}\big)^{\top}}{\sqrt{D}}\right)\mathbf{V}^{(\ell)}.
    \label{eqn:key gating}
\end{equation}

Here, $\mathrm{diag}(\mathbf{s})$ is a diagonal matrix whose entries correspond to the coefficient assigned to each token according to its group. We choose to intervene on the Key projection because scaling the keys directly modulates the pre-SoftMax logits, and therefore controls the attention distribution itself: how much probability mass each token group receives.
Hereafter, we refer to the per-group scalars in $\mathbf{s}$ as gate coefficients, and the resulting relative proportion of attention mass allocated to sink versus ordinary tokens as the attention contribution ratio.

\noindent\textbf{Experimental Setup.}\quad
We sweep all 32 self-attention layers (L0--L31) individually: for each layer, we apply the intervention to that layer alone while keeping all others at the default setting (i.e., no intervention). Within each intervened layer, the same gate coefficients are applied uniformly across all heads. We sweep the coefficients from $(0.0,\,1.0)$ to $(1.0,\,0.0)$ in steps of $0.1$ and record task accuracy for each setting. For visualization, we group the 32 layers into blocks of four and report the best-performing layer within each block.

\noindent\textbf{Token Grouping.}\quad
In Stage~1, we partition visual tokens into V-sinks ($\hat{\mathcal{I}}_{\mathrm{vit}}$) vs.\ the rest (L-sinks $+$ ordinary). Our probing shows V-sinks stably encode global information across LLM layers, and Luo \etal~\cite{luo2025sink} confirmed that V-sinks alone can suffice for global reasoning tasks. In Stage~2, we further split L-sinks from ordinary tokens for a finer local sweep.

\noindent\textbf{Target Tasks.}\quad
We evaluate on two benchmarks that span fine-grained perception and high-level reasoning. CVBench~\cite{tong2024cambrian} provides a vision-centric assessment of fine-grained capabilities such as depth, spatial relations, distance estimation, and counting, while MMStar~\cite{chen2024we} covers a broader set of task categories ranging from fine-grained perception to logical reasoning.

\begin{figure}[t]
    \vspace{-10pt}
    \centering
    \begin{subfigure}{0.48\linewidth}
        \centering
        \includegraphics[width=\linewidth]{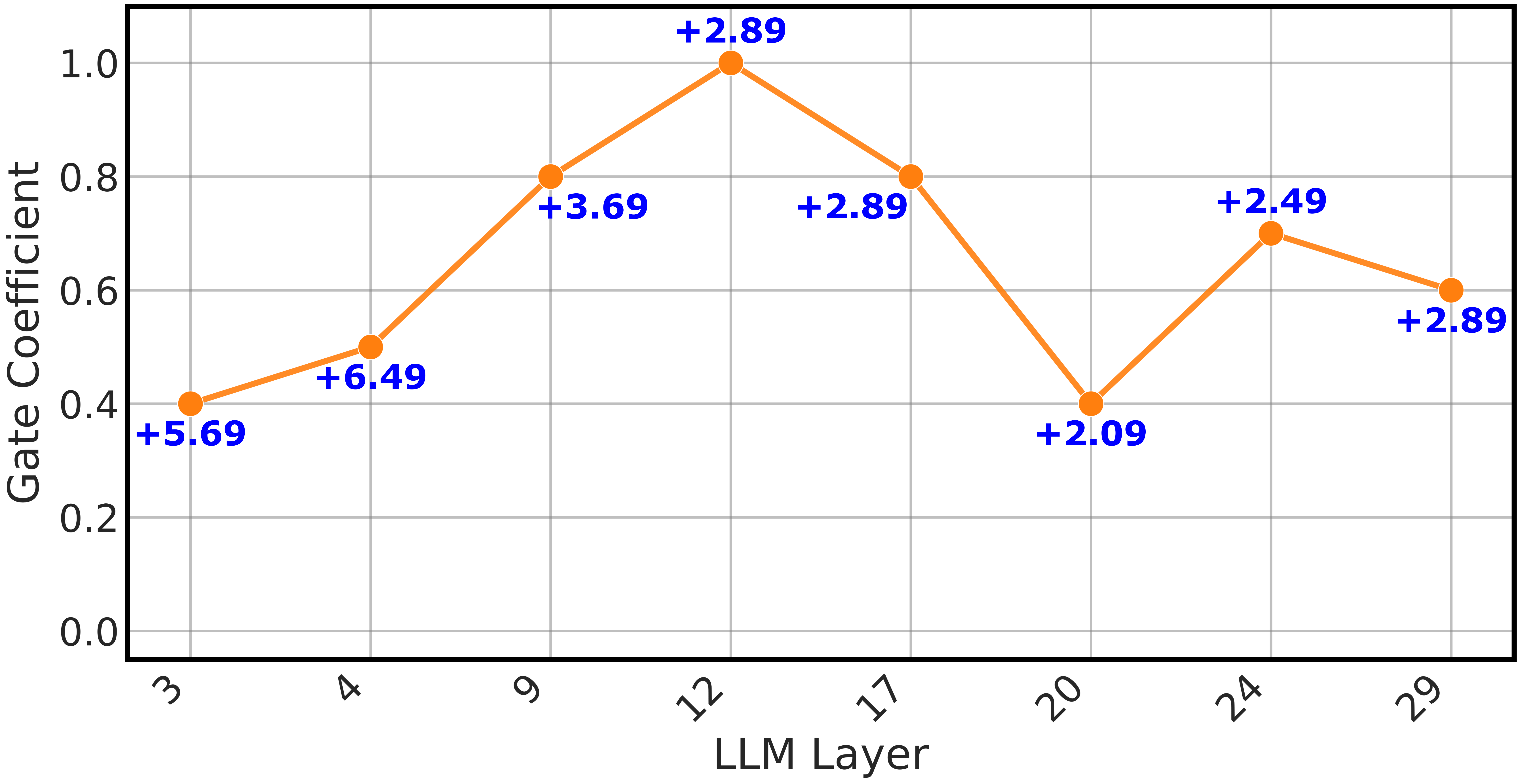}
        \caption{Fine-grained (MMStar)}
        \label{fig:fine_grained}
    \end{subfigure}\hfill
    \begin{subfigure}{0.48\linewidth}
        \centering
        \includegraphics[width=\linewidth]{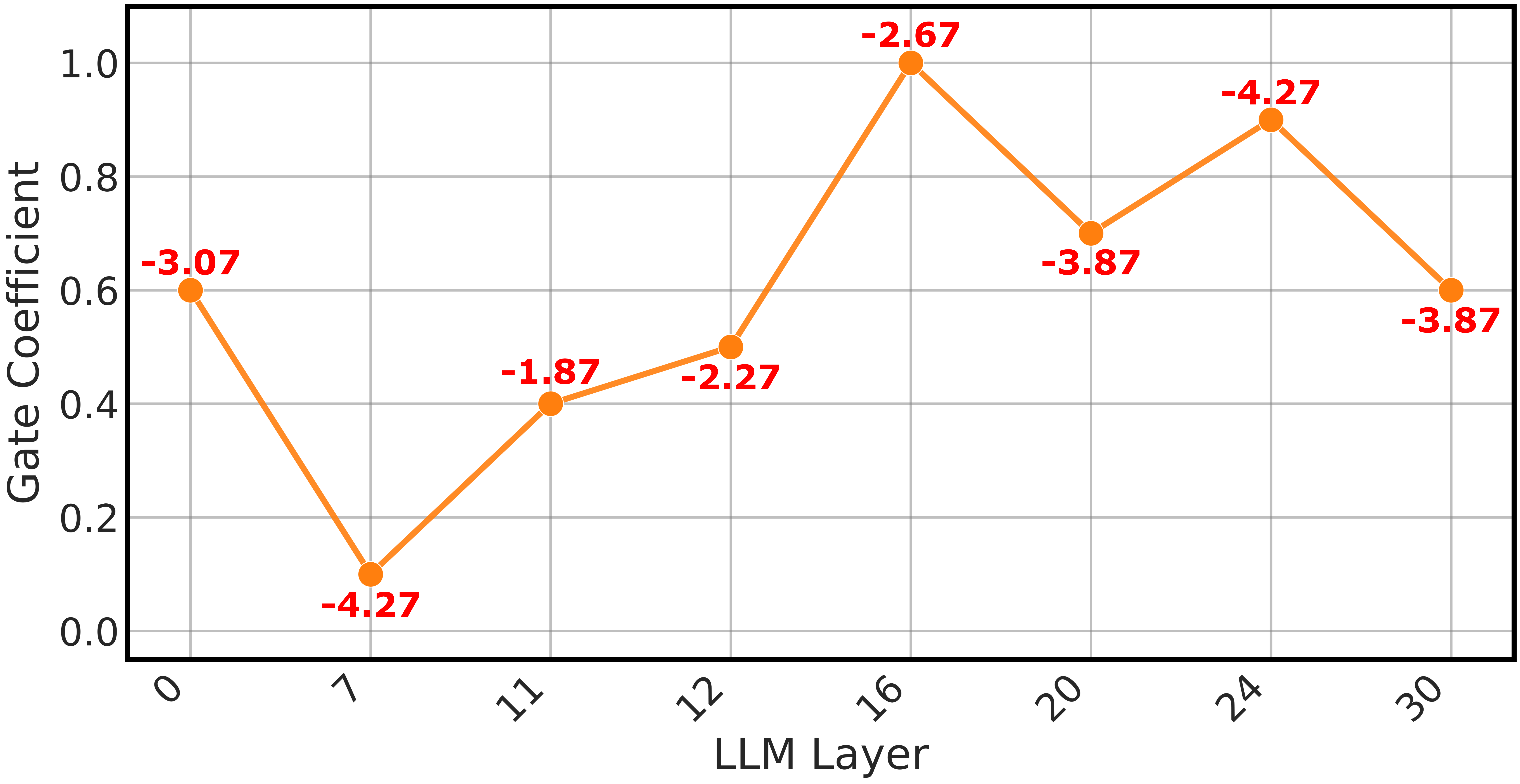}
        \caption{Coarse (MMStar)}
        \label{fig:coarse}
    \end{subfigure}\\
    \begin{subfigure}{0.48\linewidth}
        \centering
        \includegraphics[width=\linewidth]{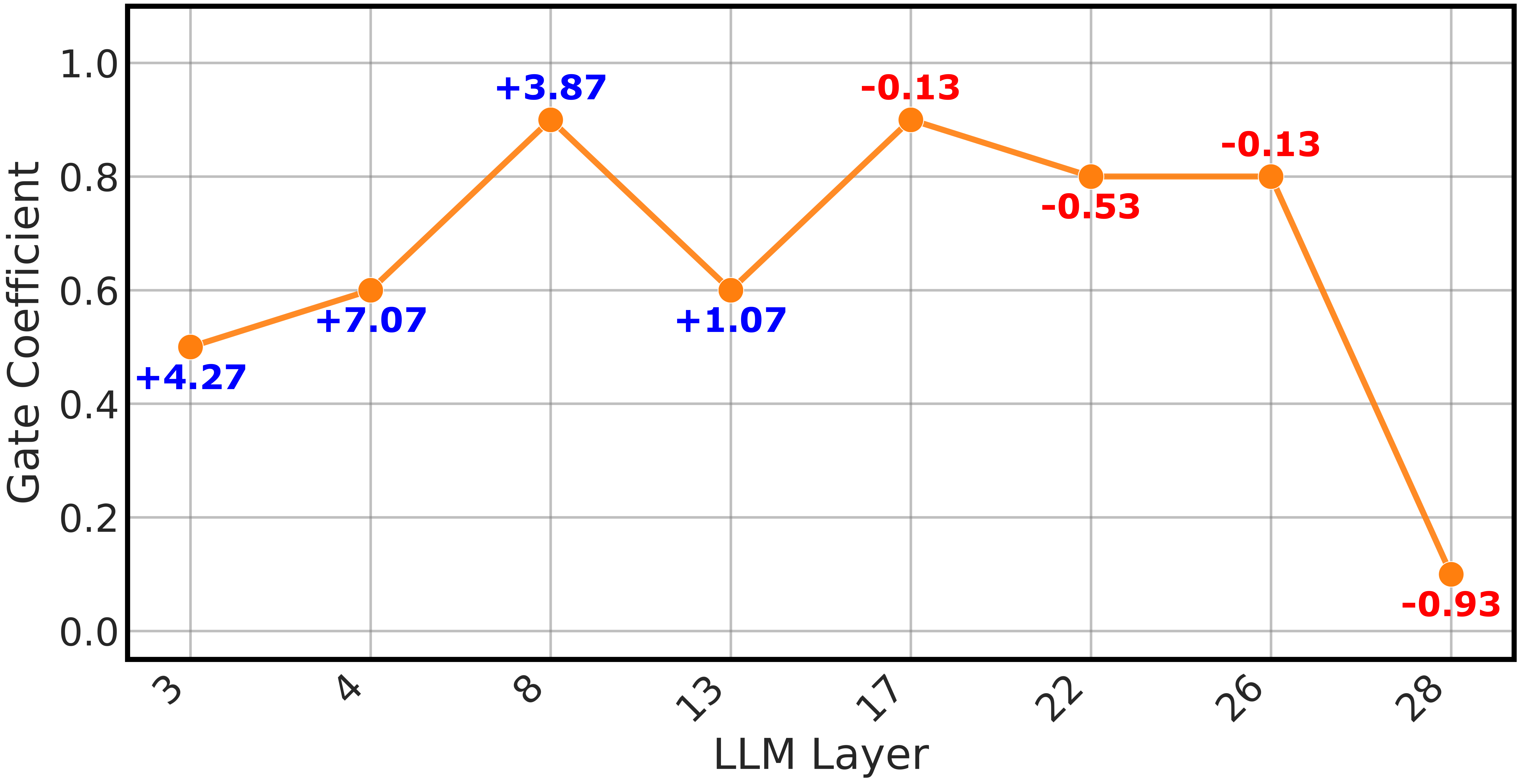}
        \caption{Logical (MMStar)}
        \label{fig:logical}
    \end{subfigure}\hfill
    \begin{subfigure}{0.48\linewidth}
        \centering
        \includegraphics[width=\linewidth]{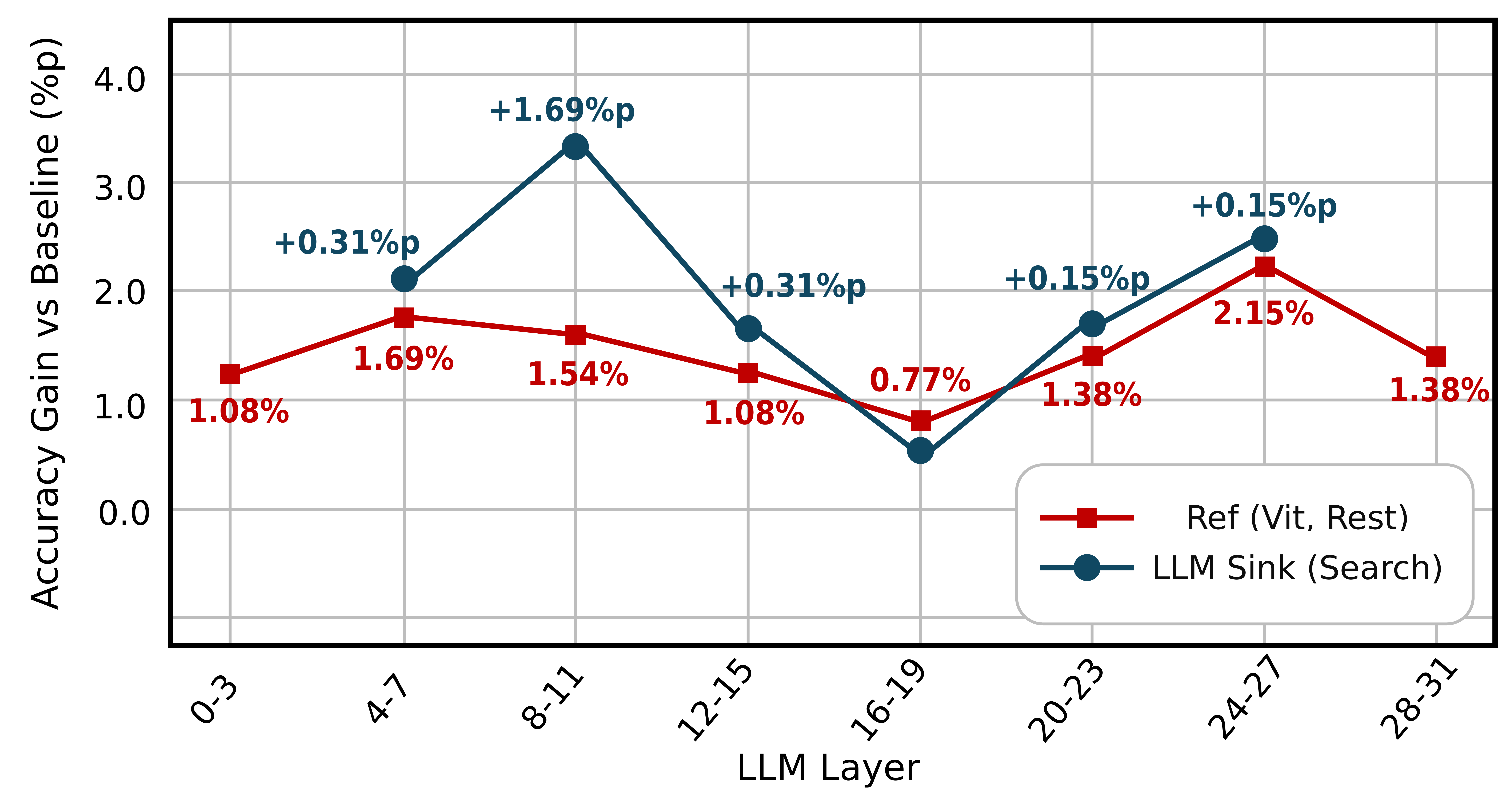}
        \caption{Relation (CVBench)}
        \label{fig:relation_llm_sink}
    \end{subfigure}
    % \vspace{-10pt}
    \caption{\textbf{Layer-wise optimal gate coefficients from key-gating intervention} (a--c)~Best sink gate per 4-layer block with accuracy change vs.\ baseline (\textcolor{blue}{positive}, \textcolor{red}{negative}). (d)~Stage-2 sweep splitting Rest into L-sinks and ordinary tokens.}
    \label{fig:gate_analysis}

    \vspace{-15pt}

\end{figure}

\noindent\textbf{Observations.}\quad Figure~\ref{fig:gate_analysis} summarizes the layer-wise optimal gate coefficients. We highlight three key findings:

\begin{itemize}
    \item \textbf{Observation 1: Task-dependency.}
          Rebalancing sink vs.\ ordinary attention does not uniformly help or hurt; its effect reverses across tasks. For fine-grained perception (Figure~\ref{fig:fine_grained}), adjusting the ratio improves accuracy at every layer block (up to +6.49\%), whereas for coarse perception (Figure~\ref{fig:coarse}), every adjustment degrades performance (up to $-$4.27\%), indicating that the default balance already suits this task and any shift disrupts it.

    \item \textbf{Observation 2: Layer-dependency.}
          Even within a single task, the effect of sink strengthening can reverse across depths: in logical reasoning (Figure~\ref{fig:logical}), early layers benefit substantially (+7.07\% at L4--L8) while late layers are hurt by the same direction (L22--L26).

    \item \textbf{Observation 3: Limited benefit of 3-group decomposition.}
          In Stage 2, splitting the residual group into L-sinks and ordinary tokens and re-sweeping around the Stage-1 optimum yields additional gains only for specific task$\times$layer combinations (Figure~\ref{fig:relation_llm_sink}), consistent with recent finding~\cite{fuhidden} that accessible information is not necessarily utilized (a detailed analysis is provided in Appendix~\ref{sec:supp_3group}).

\end{itemize}

\noindent\textbf{Takeaway.}\quad Sink tokens carry rich scene-level information that persists across all LLM layers, yet the grid search shows that exploiting it is a double-edged sword: the effect of rebalancing reverses across tasks and even across depths within a single task (Observations 1--2). A fixed coefficient cannot accommodate this variability, motivating a learned mechanism that predicts when and where to adjust sink attention.

\section{Layer-wise Sink Gating}
\label{sec:method}

\begin{figure}[t]
    \centering
    \vspace{-10pt}
    \includegraphics[width=\linewidth]{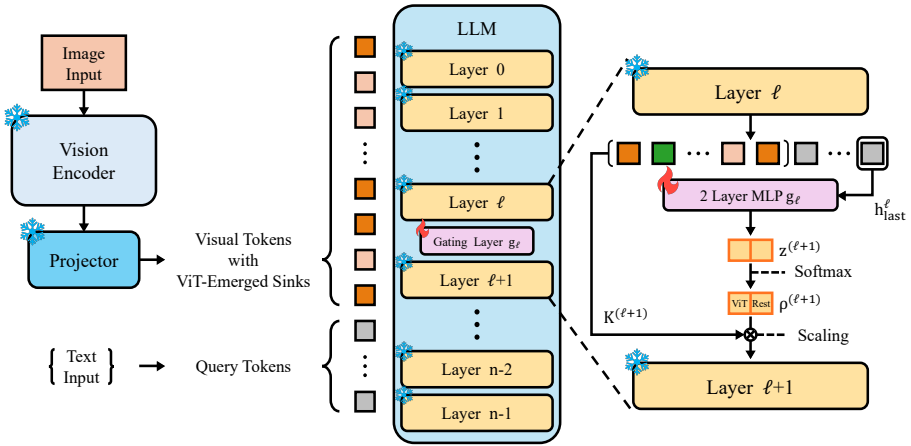}
    \vspace{-15pt}
    \caption{\textbf{Layer-wise Sink Gating (LSG).} Left: the pretrained LVLM remains frozen; a gating module (lightweight MLP) $g_\ell$ is inserted between consecutive LLM layers. Right: $g_\ell$ takes $\mathbf{h}_{\mathrm{last}}^{\ell}$ and predicts key-scaling ratios $\rho^{(\ell+1)}$ for \vsink{} vs.\ remaining visual tokens.}
    \label{fig:method}
    
    \vspace{-10pt}
    
\end{figure}

We introduce a lightweight gating module inserted between LLM layers, trained solely with the standard next-token prediction loss and no direct supervision from the grid-search oracle coefficients. The module predicts input-dependent, per-layer ratios that a fixed grid search cannot provide. Figure~\ref{fig:method} illustrates the architecture.

\subsection{Gating Signal}
\label{sec:gating_signal}

The gating module requires a per-layer input that encodes both visual and textual information. As discussed in Section~\ref{sec: hidden state stat}, the final input token progressively aggregates visual information across layers~\cite{kim2025interpreting, neotowards}, and Zhao \etal~\cite{zhao2024first} further showed that this hidden state predicts higher-order model judgments via linear probing. We therefore use $\mathbf{h}_{\mathrm{last}}^{\ell} \in \mathbb{R}^D$ as the gating signal at each layer.

\subsection{Gating Formulation}
\label{sec:gating_formulation}

 Following Observation~3 in Section~\ref{sec:grid search}, we adopt a 2-group formulation: \vsinks{} ($\hat{\mathcal{I}}_{\mathrm{vit}}$) vs.\ the remaining visual tokens ($\mathcal{I}_{\mathrm{rest}}^{\ell}$), where the rest group includes both \lsinks{} and ordinary tokens.
As shown in Figure~\ref{fig:method} (right), the gating module $g_\ell$, a lightweight 2-layer MLP, maps the gating signal at layer~$\ell$ to allocation logits for layer~$\ell{+}1$:
\begin{equation}
    \mathbf{z}^{(\ell+1)} = g_\ell\!\left(\mathbf{h}_{\mathrm{last}}^{\ell}\right) \in \mathbb{R}^{2}
\end{equation}

\noindent The attention contribution ratio between the two groups is obtained via softmax:
\begin{equation}
    \rho_{\mathrm{vit}}^{(\ell+1)},\; \rho_{\mathrm{rest}}^{(\ell+1)} = \mathrm{Softmax}\!\left(\mathbf{z}^{(\ell+1)}\right), \quad \rho_{\mathrm{vit}}^{(\ell+1)} + \rho_{\mathrm{rest}}^{(\ell+1)} = 1
    \label{eq:gate_softmax}
\end{equation}

\noindent The key vectors of each visual token $j$ at layer $\ell{+}1$ are scaled accordingly:
\begin{equation}
    \tilde{\mathbf{K}}_j^{(\ell+1)} = \rho_{g(j)}^{(\ell+1)} \cdot \mathbf{K}_j^{(\ell+1)}
    \label{eq:learned_key_gating}
\end{equation}
where $g(j) \in \{\mathrm{vit},\, \mathrm{rest}\}$ denotes the group membership of token $j$. This extends Eq.~\ref{eqn:key gating} from Section~\ref{sec:grid search} by replacing the manually swept coefficient with a learned, input-conditioned gate.
Ablations on alternative gating signals and token grouping strategies are reported in Appendix~\ref{sec:d2_ablation}.

\subsection{Training}
\label{sec:training}

All parameters of the pretrained LVLM are frozen; only the gating MLP $g$ is trained with the standard next-token prediction loss:
\begin{equation}
    \mathcal{L} = -\sum_t \log p_\theta(x_t \mid x_{<t})
\end{equation}
where $\theta$ denotes the parameters of the gating MLP. No task-specific labels or auxiliary losses are used.
Full training configuration is provided in Appendix~\ref{sec:supp_impl}.

\section{Experiment}
\label{sec:experiment}

\noindent\textbf{Experimental Setup.}\quad
We apply the gating module (Section~\ref{sec:method}) to LLaVA-1.5-7B and train it on 10K stratified samples from Cambrian-7M for 2 epochs. Each gating module contains ${\sim}$262K parameters ($4096{\to}64{\to}2$, softmax output) and is initialized to $\rho_{\mathrm{vit}}{=}0.5$, allocating equal attention to \vsink{} and remaining tokens (Section~\ref{sec:gating_formulation}). We train a separate gate at each of the 32 LLM layers independently. All results report accuracy changes (\%p) relative to the unmodified baseline on MMStar (6 sub-tasks) and CVBench (4 sub-tasks).

\begin{table}[t]
\centering
\vspace{-10pt}
\caption{Per-layer oracle (best of 11 gate values per task)
vs.\ learned gate (single layer, NTP-trained).
$\Delta$ (\%p) relative to baseline.
$n^{\geq 0}\!/\!n^{<0}$: sub-tasks with non-negative\,/\,negative
$\Delta$ out of 10 categories.}
\vspace{-10pt}

\label{tab:comparison}
\small
\setlength{\tabcolsep}{4pt}
\begin{tabular}{l l ccc ccc}
\toprule
 & & \multicolumn{3}{c}{Per-layer Oracle}
   & \multicolumn{3}{c}{Learned Gate} \\
 \cmidrule(lr){3-5} \cmidrule(lr){6-8}
Block & Layer
 & MMStar & CVBench & $n^{\geq0}\!/\!n^{<0}$
 & MMStar & CVBench & $n^{\geq0}\!/\!n^{<0}$ \\
\midrule
\multicolumn{2}{l}{Baseline}
 & 33.27 & 57.29 & & -- & -- & \\
\midrule
L0--3   & L3  & $+3.19$ & $+2.76$ & 9/1 & $+0.34$ & $+1.67$ & 8/2  \\
L4--7   & L7  & $+2.26$ & $+2.66$ & 9/1 & $+1.00$ & $+0.49$ & 8/2  \\
L8--11  & L10 & $+2.13$ & $+3.98$ & 7/3 & $+0.72$ & $+2.14$ & 8/2  \\
L12--15 & L13 & $+1.06$ & $+0.99$ & 7/3 & $-0.24$ & $+0.33$ & 6/4  \\
L16--19 & L19 & $+0.39$ & $+2.60$ & 8/2 & $+0.76$ & $+0.50$ & \textbf{10/0} \\
L20--23 & L23 & $-0.14$ & $+1.06$ & 7/3 & $+0.34$ & $+0.19$ & 9/1  \\
L24--27 & L24 & $-0.54$ & $+0.32$ & 5/5 & $+0.27$ & $-0.02$ & 7/3  \\
L28--31 & L30 & $-0.07$ & $+0.88$ & 8/2 & $+0.47$ & $+0.10$ & 8/2  \\
\bottomrule
\end{tabular}

\end{table}

\begin{figure*}[t]
    \centering
    \vspace{-5pt}
    \includegraphics[width=1.0\linewidth]{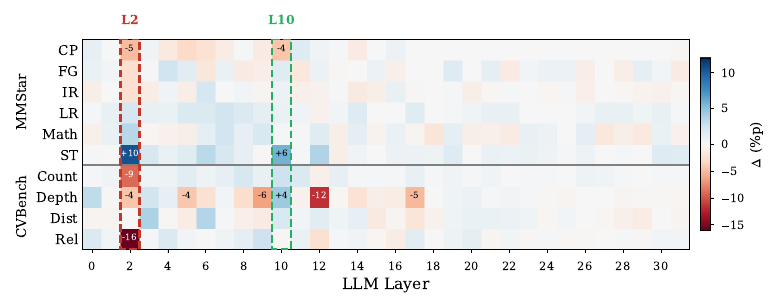}
    
    \vspace{-10pt}
    \caption{Learned gate $\Delta$ (\%p) across all 32 layers and 10 sub-tasks.
    Dashed lines mark L2 (red) and L10 (green).}
    \vspace{-10pt}
    \label{fig:heatmap}
\end{figure*}

\subsection{Single-Layer Results.}

\begin{table}[t]
\centering
\vspace{-10pt}
\caption{
  Greedy multi-layer gate stacking. Starting from the single best layer~(L10), gates are added greedily in order of marginal gain, using only the SL checkpoints without retraining. {Added}: layer appended at this step. {Active layers}: full set of stacked gates. All $\Delta$ in \%p relative to baseline. $n^{\geq0}\!/\!n^{{<}0}$: tasks with non-negative\,/\,negative $\Delta$ out of 10.
}
\vspace{-10pt}

\label{tab:greedy}
\small
\setlength{\tabcolsep}{4pt}
\begin{tabular}{c l l cc c}
\toprule
Step & Added & Active layers
  & $\Delta$MMStar & $\Delta$CVBench
  & $n^{\geq0}\!/\!n^{{<}0}$ \\
\midrule
\multicolumn{3}{l}{Baseline}
  & 0.00 & 0.00 & \\
\multicolumn{3}{l}{(single, L10)}
  & $+0.72$ & $+2.14$ & 8/2 \\
\midrule
1 & L10  & \{L10\}
  & $+0.72$ & $+2.14$ & 8/2 \\
2 & L3   & \{L3, L10\}
  & $+0.59$ & $+2.98$ & 7/3 \\
3 & L7   & \{L3, L7, L10\}
  & $+1.17$ & $+2.41$ & 7/3 \\
4 & L19  & \{L3, L7, L10, L19\}
  & $+0.74$ & $+2.47$ & 7/3 \\
5 & L6   & \{L3, L6, L7, L10, L19\}
  & $\mathbf{+1.55}$ & $\mathbf{+3.08}$ & \textbf{8/2} \\
\bottomrule
\end{tabular}

\vspace{-10pt}

\end{table}

Table~\ref{tab:comparison} reports the best single-layer result within each 4-layer block, alongside the per-layer oracle that selects the best of 11 gate values independently per task at the same layer. Seven of eight blocks yield positive gains under the learned gate; L10 achieves the largest improvement at $+0.72$ on MMStar and $+2.14$ on CVBench, while L19 maintains non-negative gains across all 10 sub-tasks. Notably, these gains emerge from NTP training on just 10K general-purpose samples, without access to task identifiers.
The oracle represents an upper bound, as it selects the best coefficient with access to ground-truth accuracy per task; the learned gate, which predicts from hidden states alone, is analyzed against this bound in Appendix~\ref{sec:supp_gap}.
For comparison, a parameter-matched LoRA shows that these gains arise from the sink-specific structure rather than from the added parameters, as analyzed in detail in Appendix~\ref{sec:d1_baseline}. We also evaluate on additional benchmarks including OCRBench and MathVista (Appendix~\ref{sec:d3_broad}).

Figure~\ref{fig:heatmap} expands the per-task accuracy changes of the learned gate to all 32 layers and 10 sub-tasks. Effective layers concentrate in L3--19, while early layers such as L2 produce large but opposing shifts across sub-tasks that hurt overall performance. We analyze the source of this layer-level variation in Section~\ref{sec:trajectory}.

\subsection{Multi-Layer Stacking}
\label{sec:stacking}

The single-layer results in Table~\ref{tab:comparison} show that multiple layers independently yield positive gains. We test whether these gains compound by stacking independently trained single-layer checkpoints without retraining: starting from L10 (the single best layer), we greedily add gates in descending order of their single-layer gain and evaluate the accumulated set jointly.

Table~\ref{tab:greedy} reports the results. The final 5-layer configuration \{L3, L6, L7, L10, L19\} achieves MMStar~$+1.55$ and CVBench~$+3.08$, exceeding L10's single-layer best of $+0.72$/$+2.14$ by a clear margin. At the same time, the stacked gains fall well below the sum of individual single-layer gains, indicating that the contributions of different layers are not independent. Adding L19 at Step~4, for instance, slightly reduces the combined score relative to Step~3, while the subsequent addition of L6 produces the largest single-step increase.
This sub-additivity reflects the distribution shift each gating module faces at inference, where hidden states have already been modified by other active gating modules not present during training.
Nonetheless, the five-layer configuration achieves a combined improvement that no single layer reaches alone, confirming that independently trained gates can be combined effectively.
The stack can involve two further choices, in how layers are selected and in
how their gates are trained. For selection, choosing the five layers from
held-out training signals alone, with no downstream-task information, yields a
stack whose performance is on par with the greedy one
(Appendix~\ref{sec:supp_leakage}). For training, the stacked gates can also be
trained jointly rather than kept independent (Appendix~\ref{sec:supp_leakage}).

% \noindent\textbf{Comparison with prior sink-handling methods.}\quad
\subsection{Comparison with prior sink-handling methods}

\begin{table*}[t]
\vspace{-10pt}
\centering
\caption{Comparison with prior sink-handling methods on LLaVA-1.5-7B.
\emph{Input-cond.}: modulation can be conditioned on the input; \emph{Layer-wise}: modulation can vary across layers. $\Delta$ (\%p) relative to the unmodified baseline
(MMStar 33.27, CVBench 57.29). $^{\dagger}$\,Single-MLP adaptation of DIYSink's CoT routing.}
\label{tab:method_comparison}
\small
\vspace{-10pt}
\renewcommand{\arraystretch}{1.15}
\begin{tabular}{l cc cc}
  \toprule
  Method & Input-cond. & Layer-wise & $\Delta$MMStar & $\Delta$CVBench \\
  \midrule
  VAR~\cite{kang2025see}                       & $\times$   & $\times$   & $-0.01$ & $+0.75$ \\
  Sink-to-the-front~\cite{luo2025sink}         & $\times$   & $\times$   & $-0.13$ & $-0.39$ \\
  DIYSink (CoT)$^{\dagger}$~\cite{luo2025sink} & \checkmark & $\times$   & $+0.25$ & $+0.41$ \\
  \midrule
  \textbf{LSG (L10)}     & \checkmark & \checkmark & $\mathbf{+0.72}$ & $\mathbf{+2.14}$ \\
  \textbf{LSG (5-layer)} & \checkmark & \checkmark & $\mathbf{+1.55}$ & $\mathbf{+3.08}$ \\
  \bottomrule
\end{tabular}
\vspace{-5pt}

\end{table*}

Table~\ref{tab:method_comparison} compares LSG against sink-related prior works
that handle \lsink{}~\cite{kang2025see} and \vsink{}~\cite{luo2025sink}. LSG at L10
alone exceeds all of them on both MMStar and CVBench, and the 5-layer stack widens
the margin to $+1.55$/$+3.08$. These gaps reflect each prior method covering only
part of Section~\ref{sec:grid search}'s (task, layer) landscape, where sink
modulation needs to both strengthen and suppress depending on the task (Obs.~1) and
the LLM layer (Obs.~2): VAR is suppress-only, sink-to-the-front is static, and
DIYSink (CoT) is input-conditioned but input-level only. In contrast, LSG
combines input-conditioned and layer-wise modulation through a lightweight learned
gate. Reproduction protocols are detailed in Appendix~\ref{sec:repro}.

\subsection{Gate Behavior Across LLM Layers}
\label{sec:trajectory}

\begin{figure}[t]
    \centering
        \centering
        \begin{subfigure}{0.48\linewidth}
            \centering
            \includegraphics[width=\linewidth]{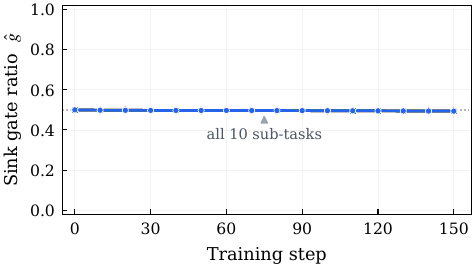}
            \caption{LLM Layer~2}
            \label{fig:traj_l2}
        \end{subfigure}\hfill
        \begin{subfigure}{0.48\linewidth}
            \centering
            \includegraphics[width=\linewidth]{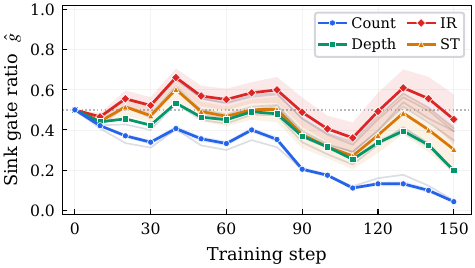}
            \caption{LLM Layer~10}
            \label{fig:traj_l10}
        \end{subfigure}
      \vspace{-5pt}
    \caption{\textbf{Learned gate trajectory during training.} Solid lines: predicted \vsink{} ratio $\rho_{\mathrm{vit}}$ averaged over evaluation samples per sub-task (shaded: $\pm 1$ std).}
    \label{fig:trajectory}

    \vspace{-10pt}
    
\end{figure}

To examine why gating effectiveness varies across layers, we record the gate's \vsink{} ratio $\rho_{\mathrm{vit}}^{(\ell)}$ per evaluation sample at each training checkpoint and aggregate by sub-task.
Figure~\ref{fig:trajectory} contrasts L2 and L10.

At L2 (left), $\rho_{\mathrm{vit}}$ remains near the $0.5$ initialization with negligible cross-task variance throughout training---the gate does not learn to differentiate inputs.
At L10 (right), the gate separates into two regimes: CVBench sub-tasks converge to low sink ratios---Count and Relation near $0.05$, Depth and Distance near $0.20$---while MMStar sub-tasks settle between $0.30$ and $0.45$. Within CVBench, these two pairs form well-separated clusters with within-task standard deviation below $0.01$.
This contrast aligns with the performance gap: L10, where the gate differentiates inputs, achieves 8 non-negative sub-tasks out of 10 (Table~\ref{tab:comparison}), whereas L2, where the gate remains near initialization, achieves only 3 non-negative sub-tasks out of 10; as Figure~\ref{fig:heatmap} shows, ST shifts by $+10$\,\%p while Rel drops by $-16$\,\%p at the same layer.

Cross-modal information transfer concentrates in layers~4--20 of LLaVA-1.5~\cite{kaduri2025s}. L2 falls below this range---at such an early depth, the final input token carries little visual information, leaving the gate no informative signal to differentiate inputs. Conversely, beyond L20, cross-modal transfer has largely concluded, so adjusting the \vsink{}-to-rest attention ratio no longer induces meaningful representational changes. This accounts for the concentration of effective layers in L3--19 visible in Figure~\ref{fig:heatmap}: the gate is effective in the layers where cross-modal transfer is still underway.
% An oracle key-gating sweep on LLaVA-OneVision-7B confirms that these task- and layer-dependent patterns persist across architectures (Appendix~\ref{sec:supp_cross_oracle}).

\subsection{Cross-Architecture Generalization}
\label{sec:arch_gen}
We examine whether our findings hold beyond LLaVA-1.5. On
LLaVA-OneVision-7B~\cite{li2025llavaonevision} (SigLIP~\cite{zhai2023sigmoid},
Qwen2-7B~\cite{qwen2}) architecture, the \vsink{}/\lsink{} taxonomy and the
oracle key-gating sweep reproduce the same task- and layer-dependent structure
(Appendix~\ref{sec:supp_cross}). On two further LLaVA variants with different
LLM and vision-encoder backbones, the oracle sweep again reproduces this
structure, and a learned single-layer LSG gives same-sign gains on both
benchmarks (Appendix~\ref{sec:cross_arch_port}), providing preliminary evidence
that LSG transfers.
\section{Conclusion}
In this work, we analyze attention sinks in LVLMs and identify two categories with distinct computational origins: ViT-emerged sinks (V-sinks) and LLM-emerged sinks (L-sinks). These sinks carry global scene priors that benefit coarse reasoning but degrade local perception. Based on this observation, we introduced Layer-wise Sink Gating, a per-layer MLP that scales the Key contributions of sink versus ordinary visual tokens. LSG is trained with next-token prediction loss alone, requires no task labels, and keeps the backbone frozen. The learned gates concentrate their effects in the mid-layers where cross-modal information transfer is active and yield gains across the evaluated benchmarks.

\vspace{-5pt}
\section*{Acknowledgements}
\vspace{-5pt}
% This work was supported in part by the National Research Foundation of Korea (RS-2024-00351212 and RS-2024-00436165) and the Institute of Information \& Communications Technology Planning \& Evaluation (IITP) (RS-2024-00509279, RS-2022-II220926, and RS-2022-II220959) funded by the Korean government (MSIT).
% This work was partially funded by the ERC (853489 – DEXIM) and the Alfried Krupp von Bohlen und Halbach Foundation, which we thank for their generous support. It was also supported by the Institute of Information \& Communications Technology Planning \& Evaluation (IITP) grant (No. RS-2026-25518317, Development of AI memory mechanism that reflects human cognitive principles) and by the National Research Foundation of Korea (NRF) grant (No. RS-2026-25495084), both funded by the Korea government (MSIT). We also thank Jaehoon Yoo for helpful discussions.
This work was partially funded by the ERC (853489 - DEXIM) and the Alfried Krupp von Bohlen und Halbach Foundation, which we thank for their generous support. It was also supported by the Institute of Information \& Communications Technology Planning \& Evaluation (IITP) grant (No. RS-2026-25518317, Development of AI memory mechanism that reflects human cognitive principles) funded by the Korea government (MSIT). This research was supported by the ANCHOR through the Seoul ANCHOR Center, funded by the Ministry of Education (MOE) and the Seoul Metropolitan Government (2026-ANCHOR-01-024-04). We also thank Jaehoon Yoo for helpful discussions.

\clearpage

\bibliographystyle{splncs04}
\bibliography{main}

\resetlinenumber[1]
\newpage
\appendix

\renewcommand{\thepage}{S\arabic{page}}
\setcounter{page}{1}
\renewcommand{\thefigure}{S\arabic{figure}}
\setcounter{figure}{0}
\renewcommand{\thetable}{S\arabic{table}}
\setcounter{table}{0}

\section*{Supplementary Materials}

\noindent
This supplement is organized as follows.

\begin{itemize}
\setlength{\itemsep}{2pt}
\setlength{\parskip}{0pt}
\item \textbf{Section~\ref{sec:supp_impl}} provides implementation details
      omitted from the main paper for space,
      including sink identification criteria,
      token-group statistics, and training configuration.
\item \textbf{Section~\ref{sec:supp_novelty}} positions our contributions
      relative to Kang~\etal~\cite{kang2025see}
      and Luo~\etal~\cite{luo2025sink}
      via a structured comparison.
\item \textbf{Section~\ref{sec:supp_oracle}} extends the oracle analysis:
      the full 32-layer$\times$10-task landscape,
      broad-optimum characterization,
      3-group vs.\ 2-group justification,
      and oracle-to-learned gap discussion.
\item \textbf{Section~\ref{sec:additional_exp}} reports additional experiments:
      a parameter-matched baseline (LoRA),
      design-choice ablations (gating signal, token grouping),
      multi-layer stacking selection and training,
      and broader benchmark evaluation.
\item \textbf{Section~\ref{sec:supp_cross}} analyzes LLaVA-OneVision-7B
      (SigLIP + Qwen2): it replicates the \vsink{}/\lsink{} taxonomy and
      repeats the oracle key-gating sweep, validating that the sink behavior
      holds on a different architecture.
\item \textbf{Section~\ref{sec:cross_arch_port}} reports the oracle
      sweep and applies LSG to additional vision-encoder and LLM backbone
      combinations.
\end{itemize}

\section{Implementation Details \& Reproducibility}
\label{sec:supp_impl}

\begin{figure}[htbp]
    \centering
    \includegraphics[width=\linewidth]{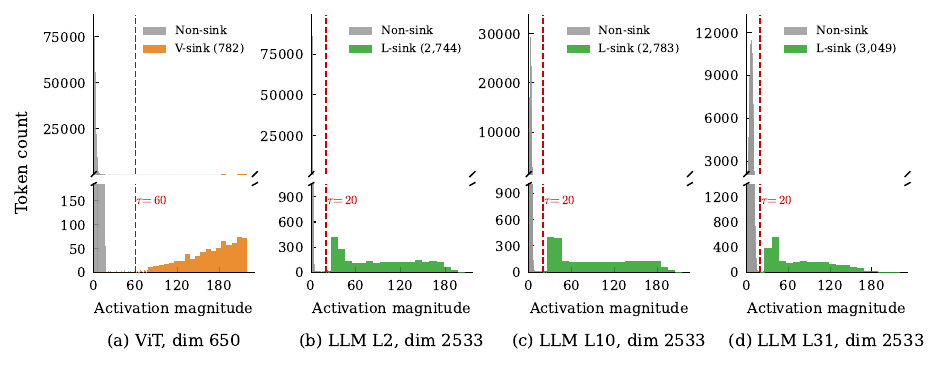}
    \caption{\textbf{Activation distribution at sink dimensions} ($300 \times 576$ visual tokens).
    (a)~ViT penultimate layer, dim~650.
    (b--d)~LLM Layers~2, 10, 31, dim~2533.
    $y$-axis broken to show both the dense bulk (top) and sparse sink tokens (bottom).
    Dashed red lines: threshold $\tau$.}
    \label{fig:threshold_dist}
\end{figure}

\subsection{Sink Identification}
\label{sec:supp_sink_id}

\noindent\textbf{Sink Dimensions.}\quad
Following Sun \etal~\cite{sun2024massive}, we identify massive activation dimensions as those whose absolute magnitudes consistently exceed all others by orders of magnitude across 300 GQA~\cite{hudson2019gqa} samples.
This yields dimension~650 for CLIP-ViT-L/14 and dimensions~1415, 2533 for LLaMA-2-7B, consistent with~\cite{sun2024massive}.
We use dimension~2533 as the primary indicator for LLM-side classification.

\noindent\textbf{Threshold $\tau$.}\quad
We set $\tau_{\mathrm{vit}} = 60$ for V-sink identification at the penultimate ViT layer (dim~650) and $\tau_{\mathrm{llm}} = 20$ for layer-wise L-sink identification in the LLM (dim~2533).
As shown in Figure~\ref{fig:threshold_dist}(a), a clear gap separates non-sink tokens (activations below ${\sim}50$) from V-sinks (above ${\sim}70$), placing $\tau_{\mathrm{vit}}$ in the empty region between the two modes.
In the LLM (Figure~\ref{fig:threshold_dist}b), the bulk of visual tokens clusters below ${\sim}10$, while L-sinks exhibit activations well above $\tau_{\mathrm{llm}}$.
The different $\tau$ values reflect the different activation scales of the ViT and LLM.
Figures~\ref{fig:threshold_dist}(c,\,d) confirm that this bimodal separation persists across LLM depths, validating a single $\tau_{\mathrm{llm}}$ for all layers.

\subsection{Token Group Statistics}
\label{sec:supp_token_stats}

\begin{figure}[t]
    \centering
    \begin{minipage}[t]{0.46\linewidth}
        \centering
        \includegraphics[width=\linewidth]{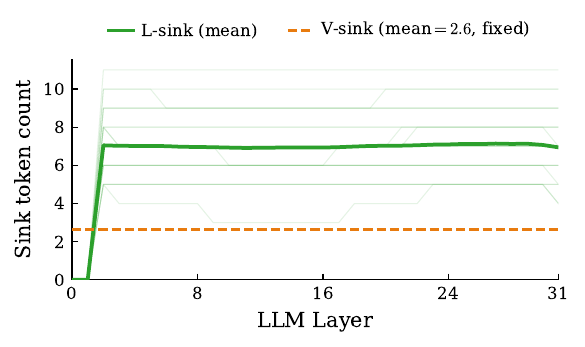}
        \captionof{figure}{Sink count across LLM layers.
        Light lines: individual samples.}
        \label{fig:sink_count}
    \end{minipage}\hfill
    \begin{minipage}[t]{0.50\linewidth}
        \centering
        \includegraphics[width=\linewidth]{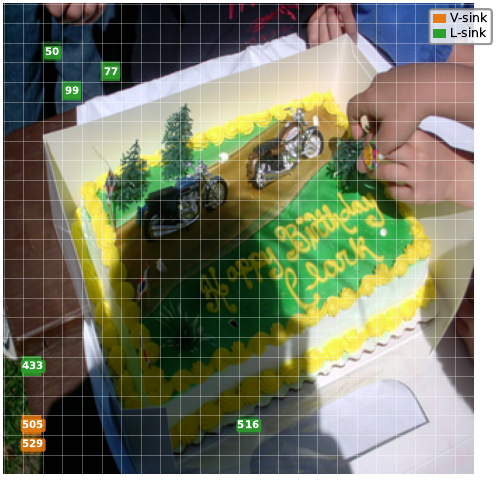}
        \captionof{figure}{Sink tokens overlaid on input image.
        \textcolor{orange}{\textbf{Orange}}: V-sinks;
        \textcolor[rgb]{0.17,0.63,0.17}{\textbf{green}}: L-sinks.}
        \label{fig:image_overlay}
    \end{minipage}
\end{figure}

We report statistics over 300 GQA samples, each yielding 576 visual tokens ($24{\times}24$ patches).
\vsinks{} average 2.6 per sample (0.5\%), fixed across all LLM layers.
\lsinks{} emerge at L2 and average ${\sim}$7 per layer (${\sim}$1.2\%), remaining stable through L31.
The remaining ${\sim}$567 tokens (98.4\%) are ordinary.
Figure~\ref{fig:sink_count} shows the layer-wise count, and Figure~\ref{fig:image_overlay} visualizes the spatial distribution on the input image from Fig.~3 of the main paper.

\subsection{Training Details}
\begin{table}[t]
  \centering
  \caption{Training configuration for LSG.}
  \label{tab:training_config}
  \small
  \begin{tabular}{@{}ll@{}}
    \toprule
    Item & Value \\
    \midrule
    Backbone          & LLaVA-1.5-7B, fully frozen \\
    Trainable params  & $g_\ell$ only (${\sim}262$K per layer) \\
    Dataset           & 10K from Cambrian-7M \\
    Stratification    & Equal across 6 categories \\
                      & (OCR, general, counting, code, math, science) \\
    Optimizer         & Adam \\
    Learning rate     & $1 \times 10^{-3}$, constant \\
    Batch size        & 128\,(8$\times$RTX\,3090, per-device 8, grad.\ accum.\ 2) \\
    Epochs            & 2 \\
    Per-layer training & Independent (32 separate runs) \\
    \bottomrule
  \end{tabular}
\end{table}

\label{sec:supp_training}

\noindent\textbf{Gating Module Architecture.}\quad
Each per-layer gating module $g_\ell$ is a two-layer MLP with LayerNorm:
$D \!\to\! 64 \!\to\! 2$, followed by Softmax (Eq.~\ref{eq:gate_softmax}).
With $D{=}4096$ (LLaMA-2-7B), the per-layer parameter count is ${\sim}262$K.
The 5-layer stacking configuration adds ${\sim}1.3$M trainable parameters in total, constituting $0.02\%$ of the 7B frozen backbone.

The Softmax output is initialized to $(\rho_{\mathrm{vit}},\,\rho_{\mathrm{rest}}) = (0.5,\,0.5)$, equivalent to applying a uniform scaling factor of $0.5$ to all visual token keys.
This mirrors the midpoint of our grid search range (Section~\ref{sec:grid search}) and provides a neutral starting point from which the gate can learn to increase or decrease sink attention in either direction.

\noindent\textbf{Training Configuration.}\quad
The entire pretrained LVLM (vision encoder, projector, and LLM) is frozen; only the gating MLP parameters are updated.
Table~\ref{tab:training_config} summarizes the setup.
Each of the 32 LLM layers is trained independently as a separate run.

The 10K training samples are drawn from Cambrian-7M~\cite{tong2024cambrian} by sampling equally from six task categories: OCR, general VQA, counting, code, math, and science (${\sim}1{,}667$ samples each).

\section{Positioning \& Novelty}
\label{sec:supp_novelty}

\begin{table}[t]
\centering
\caption{\textbf{Comparison with prior attention-sink studies in LVLMs.}}
\label{tab:novelty}
\small
\resizebox{\linewidth}{!}{
\begin{tabular}{@{}l ccc@{}}
\toprule
 & Kang~\etal~\cite{kang2025see} & Luo~\etal~\cite{luo2025sink} & Ours \\
\midrule
Sink type analyzed
  & LLM-emerged & ViT-emerged & \textbf{Both (unified)} \\
Formation mechanism
  & Massive activation & High-norm outliers & \textbf{Both origins (Sec.~3.2)} \\
Layer-wise analysis
  & --- & --- & \textbf{32-layer grid search (Obs.~1\,\&\,2)} \\
\midrule
Intervention target
  & Attention weights & Projector embeddings & Key vectors \\
Intervention granularity
  & Head-level & Model-level & Layer-level \\
Task adaptation
  & Fixed threshold & External sentence encoder & $\mathbf{h}_{\mathrm{last}}^{\ell}$ (internal, per-layer) \\
Additional module at inference
  & --- & Sentence encoder & Lightweight MLP \\
\bottomrule
\end{tabular}
}
\end{table}

Table~\ref{tab:novelty} contrasts the scope of recent attention-sink studies in LVLMs with our contributions.

Since the oracle favors suppressing sinks at some layers but strengthening them at others, even within a single task (Observation~2), a per-input, per-layer gate is necessary.
In particular, Kang~\etal~\cite{kang2025see} redistribute attention exclusively \emph{away} from sink tokens, which aligns with our findings for fine-grained perception but conflicts with reasoning tasks where the oracle favors higher sink attention ($\rho_{\mathrm{vit}}{>}0.5$) in 26 of 32 layers.
Furthermore, because the LLM's input RMSNorm re-normalizes token hidden states at every layer, scalar reweighting at the projector level~\cite{luo2025sink} has limited capacity to control the relative attention contribution of sink tokens within deeper layers; LSG instead intervenes on key vectors after normalization, where the scaling directly modulates pre-softmax attention logits.

\subsection{Reproduction of Compared Methods}
\label{sec:repro}

All results in Table~\ref{tab:method_comparison} are obtained under the same
evaluation pipeline as LSG: the same frozen LLaVA-1.5-7B checkpoint and the same
\texttt{lmms-eval} harness with deterministic decoding. We describe each compared method below.

\noindent\textbf{VAR}~\cite{kang2025see}.\quad
We run the official implementation on LLaVA-1.5-7B.\footnote{\url{https://github.com/seilk/VisAttnSink}}

\noindent\textbf{Sink-to-the-front}~\cite{luo2025sink}.\quad We follow
the paper faithfully, keeping the positional ids unchanged and only changing the
input order of the visual tokens. For \vsink{} identification,
sink-to-the-front, DIYSink, and our method all detect the same
vision-encoder-emerged tokens; we identify them with our Eq.~\ref{eq:sink_def}
criterion (described earlier), whereas Luo~\etal~\cite{luo2025sink} use a
norm-based criterion.

\noindent\textbf{DIYSink (CoT)$^{\dagger}$}~\cite{luo2025sink}.\quad DIYSink's dual-MLP projector is trained from scratch together with the LLM, which falls outside our frozen-backbone setting; we therefore compare against its component that transfers to a frozen backbone, the training-free CoT routing, applied on top of the single pretrained projector (marked $\dagger$). Among DIYSink's two routing rules we adopt CoT rather than ReW, which adds a separate text encoder.

The CoT routing asks two types of question about each input: one classifies
the image (symbolic or photographic), and the other classifies the question
(holistic or local). The two answers together determine how visual tokens are
used, in one of three modes: \texttt{mixed} (no masking), \texttt{sink\_only}
(retain only sink tokens), or \texttt{nonsink\_only} (exclude sink tokens).
To keep the decision consistent and free of answer-ordering bias, we use
word-format prompts and accept a decision only when it is stable under swapping the
order of the two options (self-consistency); the procedure is given in
Algorithm~\ref{alg:cot_routing}, and the exact prompts are shown below.

\begin{quote}\footnotesize
\begin{verbatim}
[Image-type question]
<image>
Look at the image and decide which category it belongs to.

Categories:
- symbolic: a diagram, chart, geometric figure, mathematical
  illustration, line drawing, or any image dominated by abstract or
  symbolic content with minimal local visual detail.
- photographic: a real-world photograph or naturalistic scene with
  rich visual detail.

Answer with one word: symbolic or photographic.

[Question-type question]
<image>
Given the image and the following question, decide which category the
question belongs to.

Question: {question}

Categories:
- holistic: about the overall scene, the main subject, high-level
  semantics, or content that requires understanding the image as a whole.
- local: about fine-grained details such as specific object attributes,
  spatial positions, counts, text in the image, or small regions of
  the scene.

Answer with one word: holistic or local.
\end{verbatim}
\end{quote}

\begin{algorithm}
\caption{CoT routing decision for one sample}
\label{alg:cot_routing}
\begin{algorithmic}[1]
\Require image $I$, question $q$
\Statex \textit{// each classification is run with the two answer options in both orders}
\State $a_{\mathrm{img}} \gets \textsc{ClassifyImage}(I)$ \Comment{symbolic / photographic if both orders agree, else \texttt{mixed}}
\State $a_{\mathrm{q}} \gets \textsc{ClassifyQuestion}(I, q)$ \Comment{holistic / local if both orders agree, else \texttt{mixed}}
\If{$a_{\mathrm{img}} = \text{symbolic}$ \textbf{and} $a_{\mathrm{q}} = \text{holistic}$}
    \State \Return \texttt{sink\_only}
\ElsIf{$a_{\mathrm{img}} = \text{photographic}$ \textbf{and} $a_{\mathrm{q}} = \text{local}$}
    \State \Return \texttt{nonsink\_only}
\Else
    \State \Return \texttt{mixed}
\EndIf
\end{algorithmic}
\end{algorithm}

\section{Extended Oracle Analysis}
\label{sec:supp_oracle}

\subsection{Full Oracle Landscape}
\label{sec:supp_oracle_landscape}

\begin{figure}[t]
    \centering
    \includegraphics[width=\linewidth]{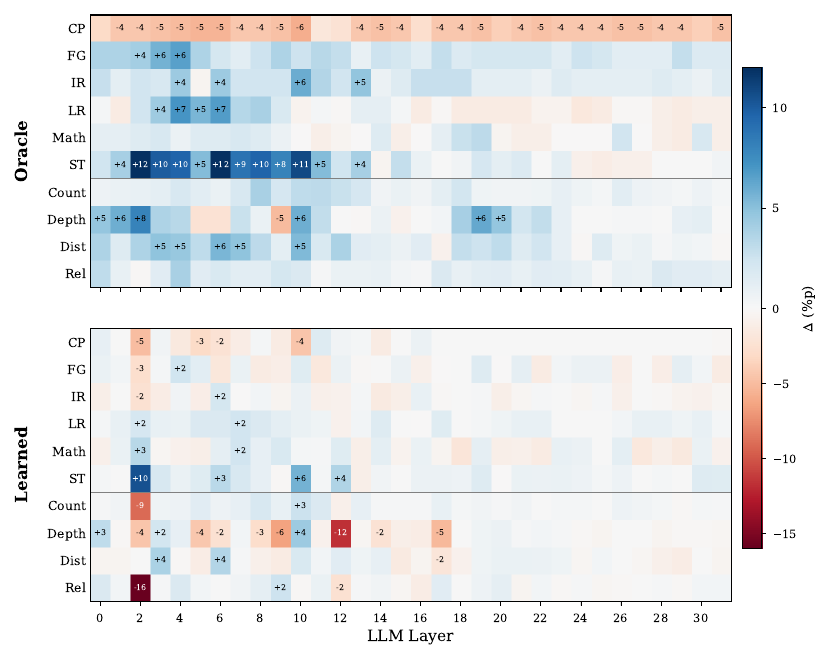}
    \caption{\textbf{Oracle (top) vs.\ Learned (bottom) gate heatmaps}
    across 32 layers and 10 sub-tasks, sharing the same color scale.
    Within L3--19, the task-wise trends are similar.}
    \label{fig:oracle_vs_learned}
\end{figure}

Figure~\ref{fig:oracle_vs_learned} presents the oracle grid search results (top)
alongside the learned-gate heatmap (bottom) across all 32 layers and 10 sub-tasks.
Each cell in the oracle row reports the best accuracy change (\%p)
obtainable by key-gating intervention at that single layer.

The oracle achieves larger absolute gains than the learned gate,
since it selects the best of 11 coefficient values independently per task.
In layers L3--19, where the learned gate yields mostly non-negative effects (Section~\ref{sec:trajectory}),
the two heatmaps show a similar task-wise trend:
ST consistently receives the strongest positive effect,
FG and LR show moderate positive effects concentrated in early-to-mid layers,
and CP is negative across all depths.
The learned gate produces smaller magnitudes overall.
We suspect this is because the NTP loss aggregates gradients across all tasks simultaneously,
which may push the gate toward conservative coefficients
that avoid harming any single task rather than toward the per-task extremes the oracle can afford.

\subsection{Broad Optimum}
\label{sec:supp_broad_optimum}

\begin{figure}[t]
    \centering
    \vspace{-5pt}
    \begin{minipage}[t]{0.48\linewidth}
        \centering
        \includegraphics[width=\linewidth]{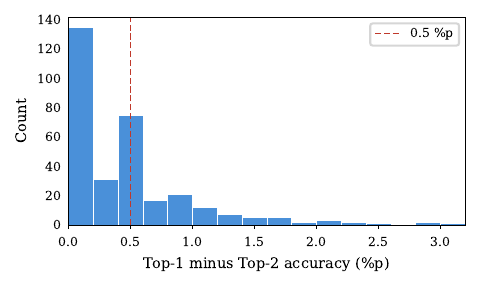}
        \captionof{figure}{\textbf{Distribution of top-1 minus top-2 accuracy gaps}
        across 320 layer$\times$task oracle settings.
        73\% fall below 0.5\%p (dashed line).}
        \label{fig:broad_optimum}
    \end{minipage}\hfill
    \begin{minipage}[t]{0.48\linewidth}
        \centering
        \includegraphics[width=\linewidth]{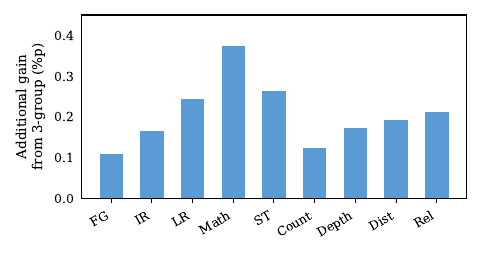}
        \captionof{figure}{\textbf{Average additional accuracy gain from 3-group over 2-group oracle},
        per sub-task. All below 0.4\%p.}
        \label{fig:3group_gain}
    \end{minipage}
\end{figure}

Across 320 layer$\times$task combinations,
73\% have a top-1 to top-2 accuracy gap below 0.5\%p,
and 42\% below 0.2\%p (Figure~\ref{fig:broad_optimum}).
In other words, for most settings the best gate value and its neighbor
yield nearly identical accuracy,
so the precise coefficient matters less than selecting the right direction
(strengthen or suppress) at the task$\times$layer combinations
where the landscape is steep (e.g., ST at L3--10, FG at L3--5).

\begin{table}[t]
  \centering
  \caption{Best single-layer accuracy gain (\%p) per sub-task: oracle vs.\ learned gate.
  L2 is excluded from both columns (see Section~\ref{sec:trajectory}).}
  \label{tab:oracle_learned_gap}
  \small
  \begin{tabular}{@{}lccccc@{}}
    \toprule
    & Oracle & Layer & Learned & Layer & Recovery \\
    \midrule
    FG    &  +6.49 & L4  &  +2.37 & L4  & 36.5\% \\
    IR    &  +5.94 & L10 &  +2.15 & L6  & 36.2\% \\
    LR    &  +7.07 & L4  &  +2.32 & L7  & 32.8\% \\
    Math  &  +3.11 & L19 &  +2.26 & L7  & 72.7\% \\
    ST    & +12.02 & L6  &  +5.66 & L10 & 47.1\% \\
    \midrule
    Count &  +3.93 & L8  &  +2.54 & L10 & 64.5\% \\
    Depth &  +6.17 & L19 &  +4.33 & L10 & 70.2\% \\
    Dist  &  +5.50 & L6  &  +3.83 & L3  & 69.7\% \\
    Rel   &  +3.85 & L4  &  +2.46 & L9  & 64.0\% \\
    \bottomrule
  \end{tabular}
\end{table}

\subsection{Why V-sink vs.\ Rest: 3-Group Oracle Analysis}
\label{sec:supp_3group}

As noted in Section~\ref{sec:supp_novelty}, the learned gate uses a 2-group partition
(V-sinks vs.\ the rest) despite our analysis identifying three token groups.
Here we provide the quantitative rationale.

\noindent\textbf{Setup.}\quad
Using the hierarchical sweep described in Section~3.4 of the main paper,
we fix the V-sink coefficient at its 2-group optimum and further split the ``rest'' group
into L-sinks and ordinary tokens, sweeping their coefficients independently.
This yields a 3-group oracle for 232 layer$\times$task combinations
(9 sub-tasks; coarse perception is excluded as no positive gain exists in the 2-group oracle).

\noindent\textbf{The 2-group gate recovers most of the 3-group gain.}\quad
Despite the 3-group oracle assigning L-sinks a different coefficient than ordinary tokens
in 82\% of layer$\times$task settings,
the 2-group gate still recovers over 90\% of the 3-group accuracy gain
in 69\% of all settings.
The average additional gain from the third degree of freedom is only 0.195\%p
(Figure~\ref{fig:3group_gain}).

\noindent\textbf{Information redundancy.}\quad
This is consistent with our probing results (Section~3.3):
both V-sinks and L-sinks encode similar scene-level global attributes (count, color, shape, size).
Since adjusting V-sink attention already modulates the model's access to global information,
an independent L-sink coefficient has limited room to provide additional benefit.

\subsection{Oracle-to-Learned Gap}
\label{sec:supp_gap}

Table~\ref{tab:oracle_learned_gap} reports the best single-layer gain for each sub-task.
Following Section~\ref{sec:trajectory}, L2 is excluded from both columns,
as the learned gate at L2 does not differentiate inputs.
Excluding L2 provides a fair comparison between the oracle and learned settings.

\noindent\textbf{Interpreting the gap.}\quad
The oracle selects the best of 11 coefficient values independently per task,
so it represents a per-task upper bound rather than a target the learned gate is expected to match.
The learned gate predicts a single coefficient per input from $\mathbf{h}_{\mathrm{last}}^{\ell}$,
trained with NTP loss on 10K general-purpose samples without task identifiers.

\noindent\textbf{Directional agreement.}\quad
The grid search (Section~3.4) identified task- and layer-dependent trade-offs
through exhaustive sweeps.
The learned gate, trained with NTP loss alone
and without access to oracle coefficients or task labels,
reproduces this pattern:
it predicts lower sink ratios for fine-grained inputs
and higher ratios for reasoning inputs (Figure~\ref{fig:trajectory}),
and its gains concentrate in L3--19 (Figure~\ref{fig:oracle_vs_learned}),
largely overlapping with the cross-modal transfer range (layers 4--20) identified by~\cite{kaduri2025s}.
This confirms that $\mathbf{h}_{\mathrm{last}}^{\ell}$ carries sufficient signal
for the gate to become task-separable under NTP training alone,
without requiring oracle coefficients or task labels.

\section{Additional Experiments}
\label{sec:additional_exp}

We conduct additional experiments to (i) compare LSG against a generic parameter-efficient baseline (Sec.~\ref{sec:d1_baseline}), (ii) ablate key design choices in the gating mechanism (Sec.~\ref{sec:d2_ablation}), (iii) study multi-layer stacking, including layer selection from training signals and independent versus joint gate training (Sec.~\ref{sec:supp_leakage}), and (iv) evaluate on bro
ader benchmarks beyond those used in the main paper (Sec.~\ref{sec:d3_broad}).
Unless otherwise stated, all learned variants are trained on a 10K stratified sample from Cambrian-7M with a frozen backbone.

\subsection{Baseline Comparison}
\label{sec:d1_baseline}

A natural question is whether the improvements from LSG arise from the sink-specific gating structure identified in our analysis, or from the additional parameters alone.
To test this, we compare LSG against LoRA~\cite{hu2022lora} applied to a single attention layer (L10), matching the parameter budget as closely as possible.

\noindent\textbf{Setup.}\quad
We consider two conditions beyond the unmodified baseline: (1) \textit{LoRA}: rank-8 LoRA on all four projections ($\mathbf{W}_q, \mathbf{W}_k, \mathbf{W}_v, \mathbf{W}_o$) at L10, totaling $4 \times 2 \times 4096 \times 8 = 262$K parameters; (2) \textit{LSG}: our gating mechanism at L10 with 262K parameters.
The parameter counts coincide, providing a controlled comparison.
Both conditions use the same training configuration (10K samples, 2 epochs, NTP loss).

\noindent\textbf{Results.}\quad
Table~\ref{tab:d1_baseline} reports deltas against the unmodified baseline.
LSG and LoRA show comparable MMStar gains ($+0.72$ vs.\ $+0.70$\,pp), but their effect profiles diverge on spatial reasoning: LSG improves CVBench by $+2.14$ while LoRA degrades it by $-5.30$.
Despite identical parameter budgets, the two methods produce qualitatively different trade-off patterns, confirming that LSG's gains stem from the sink-specific structure identified in our analysis rather than from additional parameters alone.

\begin{table}[t]
\centering
\caption{\textbf{Baseline comparison at L10.} $\Delta$ values are relative to the unmodified LLaVA-1.5-7B baseline. Both methods use 262K trainable parameters with the same 10K NTP setup. LoRA uses rank-8 on $\{q,k,v,o\}$ projections.}
\label{tab:d1_baseline}
\small
\begin{tabular}{l c c c}
\toprule
Method & Params & $\Delta$MMStar & $\Delta$CVBench \\
\midrule
Baseline & 0 & 33.27 & 57.29 \\
\midrule
LoRA rank-8 & 262K & $+0.70$ & $-5.30$ \\
\textbf{LSG (ours)} & 262K & $\mathbf{+0.72}$ & $\mathbf{+2.14}$ \\
\bottomrule
\end{tabular}
\end{table}

\subsection{Design Choice Ablations}
\label{sec:d2_ablation}

LSG involves two design choices: (A) which hidden state to use as the gating signal, and (B) how to partition tokens into groups.
All ablations are conducted at L10, the best single-layer configuration from the main paper, and evaluated on MMStar and CVBench.
Table~\ref{tab:d2_ablation} reports these design-choice ablations at L10.

\begin{table}[t]
\centering
\caption{\textbf{Design choice ablations at L10.} The first row in each category corresponds to the default LSG configuration. $\Delta$ values are relative to the unmodified baseline. All variants use the same training setup (10K samples, 2 epochs). The $n{\geq}0$\,/\,$n{<}0$ column counts non-negative vs.\ negative subtasks out of 10 (MMStar $\times$ 6 + CVBench $\times$ 4).}
\label{tab:d2_ablation}
\small
\resizebox{\linewidth}{!}{
\begin{tabular}{l l c r r c}
\toprule
Category & Variant & Params & $\Delta$MMStar & $\Delta$CVBench & $n{\geq}0$ / $n{<}0$ \\
\midrule
    & Baseline & 0 & 33.27  & 57.29 & --- \\
    & \textbf{LSG (ours)} & \textbf{262K} & $\mathbf{+0.72}$ & $\mathbf{+2.14}$ & \textbf{8\,/\,2} \\
\midrule
\multirow{3}{*}{\textbf{A. Gating signal}} 
  & $\mathbf{h}_{\text{last}}$ (ours)           & 262K & $+0.72$ & $+2.14$ & 8\,/\,2 \\
  & Mean-pool (all tokens)   & 262K & $-0.28$ & $-0.07$ & 5\,/\,5 \\
  & Mean-pool (visual only)  & 262K & $-0.30$ & $-0.15$ & 3\,/\,7 \\
\midrule
\multirow{3}{*}{\textbf{B. Token grouping}} 
  & V-sink vs.\ rest (ours) & 262K & $+0.72$ & $+2.14$ & 8\,/\,2 \\
  & L-sink vs.\ rest        & 262K & $+1.07$ & $+1.75$ & 6\,/\,4 \\
  & 3-group (V / L / ord)   & ${\sim}$262K & $+0.98$ & $+2.96$ & 8\,/\,2 \\
\bottomrule
\end{tabular}
}
\end{table}

\noindent\textbf{A. Gating signal.}\quad
We compare three sources for the gating signal: $\mathbf{h}_{\text{last}}$ (the hidden state of the last text token, our default), a mean-pool over all input tokens, and a mean-pool restricted to visual tokens.
$\mathbf{h}_{\text{last}}$ is the only variant that yields consistent gains on both MMStar ($+0.72$\,pp) and CVBench ($+2.14$), with 8 out of 10 subtasks showing non-negative change.
Both mean-pooled variants produce near-zero aggregate deltas, but the subtask breakdown reveals a more concerning pattern: mean-pool over visual tokens degrades 7 out of 10 subtasks, while mean-pool over all tokens shows no clear direction (5\,/\,5).
This is consistent with how autoregressive LLMs aggregate context into the final token~\cite{neotowards, kim2025interpreting}; mean-pooling dilutes this concentrated signal into a position-averaged representation that lacks query specificity.

\noindent\textbf{B. Token grouping.}\quad
We compare our default 2-group partition (V-sink vs.\ rest) against two alternatives: (i) grouping by LLM-emerged sinks (L-sink vs.\ rest), and (ii) a 3-group formulation that assigns separate softmax-normalized ratios to V-sinks, L-sinks, and ordinary tokens.\footnote{The 3-group formulation uses a 3-way softmax, jointly optimizing all three ratios end-to-end. This differs from the hierarchical oracle sweep in Sec.~3.4, which first fixes the V-sink ratio and then subdivides the remainder.}

All three groupings produce positive gains on both MMStar and CVBench, with no single variant dominating across both metrics: L-sink leads on MMStar ($+1.07$), 3-group leads on CVBench ($+2.96$), and V-sink falls between ($+0.72$, $+2.14$).
The similar overall gains across groupings suggest that, at L10, separating \textit{any} sink set from ordinary tokens modulates access to the same global information.
This is consistent with the probing results at this layer (Sec.~3.3), and with Observation~3, where the 2-group partition recovered most of the 3-group oracle gain.
We adopt V-sink vs.\ rest as the default given its subtask consistency (8\,/\,2) and the simplicity of a fixed token partition.

\noindent\textbf{Epoch stability.}\quad
As a partial stability check in lieu of full multi-seed evaluation, we compare epoch-1 and epoch-2 checkpoints across all single-layer configurations.
Across 320 layer$\times$task combinations from the 32-layer grid search, only 5.3\% of configurations exhibit a sign flip between epochs.
The top-performing layers (L3, L7, L10, L19) remain within the top 8 in both epochs, and L19 converges to nearly identical gate values across checkpoints.
We report epoch-1 results throughout as the less overfit checkpoint; the consistency across epochs provides evidence for training stability despite the limited 10K sample size.

\subsection{Multi-Layer Stacking: Selection and Training}
\label{sec:supp_leakage}

The greedy stack in Section~\ref{sec:stacking} builds a multi-layer configuration
by ordering layers by their single-layer benchmark gains and stacking independently
trained gates. This involves two choices: how the layers are selected, and how their
gates are trained. We report an alternative for each, summarized in
Table~\ref{tab:leakage_free}.

\noindent\textbf{Selection from training signals.}\quad
We give an alternative to the greedy order that ranks layers using only signals
computable from held-out training data, with no access to downstream-task
information. We first discard layers whose gate stays near initialization and therefore does not differentiate inputs (consistent with Section~\ref{sec:trajectory}). For each remaining layer $L$ we compute two
quantities over the same six task categories used for training
(Appendix~\ref{sec:supp_training}). The first, $w(L)$, is the standard deviation of
the predicted $\rho_{\mathrm{vit}}^{(L)}$ within each task category, averaged over
categories. A small $w(L)$ means the gate responds consistently to inputs of the
same category, which we take as a sign that the gate has learned a meaningful,
task-dependent allocation rather than a near-uniform one. The second, $g(L)$, is the
held-out next-token prediction loss of the model with the layer-$L$ gate active; a
lower $g(L)$ reflects a gate that has been fit well on the held-out data. We then
standardize each signal across the remaining layers and sum,
\begin{equation}
S(L) = z\!\left(w(L)\right) + z\!\left(g(L)\right),
\qquad
z(x_L) = \frac{x_L - \operatorname{mean}_{L'} x_{L'}}{\operatorname{std}_{L'} x_{L'}},
\label{eq:leakage_score}
\end{equation}
taking the top-$K$ layers by lowest $S$.
At $K{=}5$, the selected stack $\{$L3, L4, L6, L10, L26$\}$ performs on par with the
greedy one (Table~\ref{tab:leakage_free}), with no access to downstream-task
information.

\noindent\textbf{Training the stacked gates.}\quad
Joint training achieves a higher CVBench gain ($+3.67$ vs.\ $+3.08$) than greedy
stacking, suggesting a benefit from cross-layer co-adaptation. However, MMStar is
lower ($+0.72$ vs.\ $+1.55$), likely because the $5\times$ larger parameter space
(1.3M vs.\ 262K) relative to the fixed 10K training set increases overfitting risk.
Scaling joint training to larger datasets may close this gap; we leave this to future
work.

\begin{table}[t]
\centering
\caption{Multi-layer stacking variants on LLaVA-1.5-7B. Starting from the
greedy reference, the \emph{Training-signal selection} block chooses layers without
downstream-task information ($k{=}1,2,5$), and \emph{Joint training} trains the five
stacked gates together. $\Delta$ (\%p) relative to baseline.}
\label{tab:leakage_free}
\small
\setlength{\tabcolsep}{4pt}
\begin{tabular}{llcc}
\toprule
 & Layers & $\Delta$MMStar & $\Delta$CVBench \\
\midrule
Greedy (reference)         & \{3,6,7,10,19\} & $+1.55$ & $+3.08$ \\
\midrule
\multicolumn{4}{l}{\textit{Training-signal selection}} \\
  & \{10\}            & $+0.72$ & $+2.14$ \\
  & \{3,10\}          & $+0.59$ & $+2.98$ \\
  & \{3,4,6,10,26\}   & $+1.42$ & $+3.11$ \\
\midrule
Joint training             & \{3,6,7,10,19\} & $+0.72$ & $+3.67$ \\
\bottomrule
\end{tabular}
\end{table}

\subsection{Broader Benchmark}
\label{sec:d3_broad}
 
\begin{table}[t]
  \centering
  \caption{\textbf{Broader benchmark evaluation.}
    $\Delta$ from unmodified LLaVA-1.5-7B baseline.
    $\uparrow$\,/\,$\downarrow$: benchmarks with positive
    vs.\ negative change.}
  \label{tab:broader_bench}
  \small
  \setlength{\tabcolsep}{5pt}
  \begin{tabular}{@{}ll r rrr@{}}
    \toprule
    \textbf{Domain} & \textbf{Benchmark} & \textbf{Base}
      & $\boldsymbol{\Delta}$\textbf{L2}
      & $\boldsymbol{\Delta}$\textbf{L10}
      & $\boldsymbol{\Delta}$\textbf{L19} \\
    \midrule
    Spatial     & CVBench      &  57.3  & $-$7.3   & \textbf{+2.1}  & +0.5   \\
    Multimodal  & MMStar       &  33.3  & +0.9     & +0.7           & +0.8   \\
    OCR         & OCRBench     & 312    & $-$1.0   & +2.0           & \textbf{+3.0}   \\
    Math        & MathVista    &  22.6  & $-$1.2   & $-$0.6         & +0.1   \\
    \midrule
    \multicolumn{3}{@{}l}{$\uparrow$\,/\,$\downarrow$}
      & 1\,/\,3 & 3\,/\,1 & \textbf{4\,/\,0} \\
    \bottomrule
  \end{tabular}
\end{table}

The main paper evaluates LSG on MMStar and CVBench.
Here we extend the evaluation to two additional benchmarks
covering complementary capability domains:
OCRBench~\cite{liu2024ocrbench} (text recognition) and
MathVista~\cite{lu2024mathvista} (mathematical reasoning).
We compare L2 (negative control),
L10 (best single-layer), and
L19 (safest single-layer; 10/10 non-negative subtasks
in the main evaluation).

Table~\ref{tab:broader_bench} shows the results.
L2 degrades 3 of 4 benchmarks,
confirming that early-layer gating without an informative signal
is broadly disruptive (consistent with Section~\ref{sec:trajectory}).
L19 improves all four benchmarks,
including OCRBench ($+3.0$) and MathVista ($+0.1$).
L10 achieves larger gains on CVBench ($+2.1$) and OCRBench ($+2.0$)
but degrades MathVista ($-0.6$).

\begin{table}[t]
\centering
\caption{\textbf{Architectural comparison.}}
\label{tab:arch_comparison}
\small
\begin{tabular}{@{}lcc@{}}
\toprule
 & LLaVA-1.5-7B & LLaVA-OV-7B \\
\midrule
Vision encoder & CLIP-ViT-L/14 & SigLIP-SO400M \\
Resolution     & Fixed 336\,px   & Dynamic (AnyRes) \\
LLM            & LLaMA-2-7B (32L, $D{=}4096$) & Qwen2-7B (28L, $D{=}3584$) \\
Visual tokens  & 576 (fixed)   & 729 base + varies \\
\midrule
$\mathcal{D}_{\mathrm{sink}}$ (ViT) & \{650\}          & \{1076\} \\
$\mathcal{D}_{\mathrm{sink}}$ (LLM) & \{1415, 2533\} & \{458, 2570\} \\
$\tau_{\mathrm{vit}}$\;/\;$\tau_{\mathrm{llm}}$ & 60\;/\;20 & 40\;/\;50 \\
\bottomrule
\end{tabular}
\end{table}

\section{Sink Analysis on LLaVA-OneVision}
\label{sec:supp_cross}

% \begin{wrapfigure}{r}{1.0\textwidth}
\begin{figure}
  % \vspace{-30pt}
  \centering
  \includegraphics[width=\linewidth]{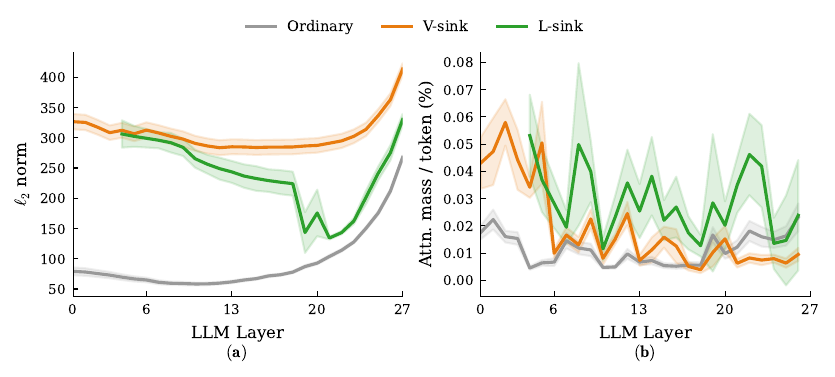}
  \caption{\textbf{Layer-wise salience on LLaVA-OneVision-7B}
  (cf.\ Fig.~2).
  (a)~$\ell_2$ norms;
  (b)~per-token attention mass
  from the last input token.}
  \label{fig:ov_salience}
\end{figure}

All analyses in the main paper use LLaVA-1.5-7B (CLIP-ViT-L/14 + LLaMA-2-7B).
To test whether the V-sink/L-sink taxonomy generalizes,
we replicate the core analyses on LLaVA-OneVision-7B~\cite{li2025llavaonevision},
whose architectural differences are summarized in Table~\ref{tab:arch_comparison}.

\subsection{Experimental Setup}
\label{sec:supp_cross_setup}

All analyses use 300~GQA~\cite{hudson2019gqa} samples,
matching the protocol for LLaVA-1.5.
LLaVA-OneVision employs AnyRes,
splitting each image into a base crop and multiple sub-crops,
each independently encoded by SigLIP.
We identify V-sinks across all crops (base and sub-crops)
by applying Eq.~(\ref{eq:sink_def})
at dim~1076 of each SigLIP patch output
(thresholds listed in Table~\ref{tab:arch_comparison}).

\begin{figure}[t]
    \centering
    \includegraphics[width=\linewidth]{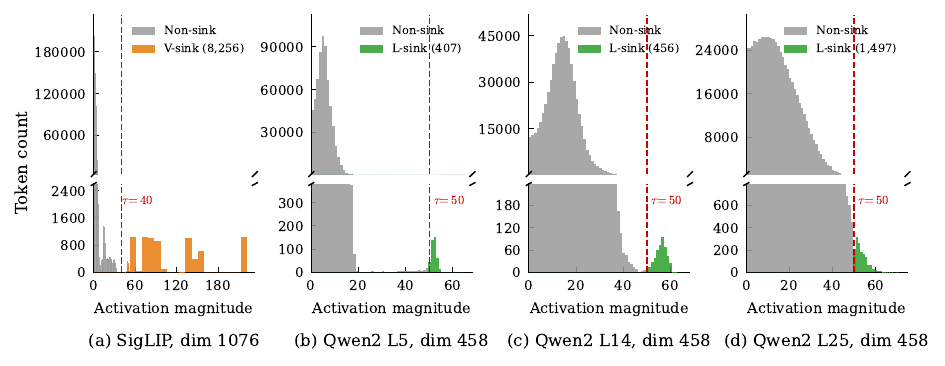}
    \caption{\textbf{Activation distribution at sink dimensions
    on LLaVA-OneVision-7B} (cf.\ Figure~\ref{fig:threshold_dist}).
    (a)~SigLIP dim~1076 ($\tau_{\mathrm{vit}}{=}40$).
    (b--d)~Qwen2 dim~458 at Layers~5, 14, 25 ($\tau_{\mathrm{llm}}{=}50$);
    the gap between bulk and sink narrows at deeper layers.}
    \label{fig:ov_threshold}
\end{figure}

\noindent\textbf{Sink dimensions and thresholds.}\quad
Following Sun~\etal~\cite{sun2024massive},
we identify massive-activation dimensions
as those whose absolute magnitudes consistently exceed
all others by orders of magnitude.
For SigLIP-SO400M~\cite{zhai2023sigmoid}, dimension~1076 serves as the primary V-sink indicator.
As shown in Figure~\ref{fig:ov_threshold}(a),
the bulk distribution concentrates below ${\sim}10$
and sink tokens appear sparsely above ${\sim}40$,
with a wide gap (${\sim}10$ to ${\sim}40$) in between;
we set $\tau_{\mathrm{vit}}{=}40$.
For Qwen2-7B~\cite{qwen2}, dimensions~458 and 2570 serve as the LLM sink dimensions,
consistent with Kang~\etal~\cite{kang2025see};
we use dim~458 as the primary indicator for L-sink classification.
At early stable layers (e.g., L5; Figure~\ref{fig:ov_threshold}b),
the bulk stays below ${\sim}15$ and L-sinks appear above ${\sim}50$,
providing a wide gap for $\tau_{\mathrm{llm}}{=}50$.
At deeper layers,
Qwen2's growing residual stream norm
shifts the bulk distribution rightward,
progressively narrowing the gap.
At L14 (Figure~\ref{fig:ov_threshold}c),
the bulk extends to ${\sim}40$ and the gap remains sufficient.
At L25 (Figure~\ref{fig:ov_threshold}d),
the bulk reaches ${\sim}45$,
leaving insufficient margin for reliable separation;
the L-sink count at this layer (1{,}497)
is notably higher than at L5 (407) or L14 (456),
confirming that the fixed threshold captures many borderline tokens
at this depth.
We apply Eq.~(\ref{eq:sink_def})
with raw absolute activation magnitudes throughout.

\noindent\textbf{Token statistics.}\quad
Over 300~GQA samples,
\vsinks{} average ${\sim}$20 per image
across all AnyRes crops combined (base crop $+$ sub-crops).
\lsinks{} emerge at L4 and average ${\sim}$1--5 per layer
within L4--L18, where the bimodal gap at dim~458
is wide enough for stable identification.
Starting from L19, L-sink counts spike sharply
(peak ${\sim}$50--100 at L20--L22;
Figure~\ref{fig:ov_sink_count}),
coinciding with the residual norm growth
discussed above.
This is a threshold artifact rather than genuine L-sink emergence.
Counts decrease at L25--L26 (${\sim}$5--8 per layer);
L27 shows a modest increase but remains
well below the L19--L24 peak.

\subsection{Sink Taxonomy Replication}
\label{sec:supp_cross_taxonomy}

\noindent\textbf{Salience pattern (Figure~\ref{fig:ov_salience}).}\quad
\vsinks{} remain above ordinary tokens in $\ell_2$ norm
across all 28~layers (panel~a),
though the gap narrows in mid-layers (L13--L18)
before widening again as residual norms grow.
\vsinks{} also receive disproportionate attention mass
from the last input token throughout (panel~b).
\lsinks{} similarly remain above ordinary tokens
in both norm and attention mass within L4--L27.
Overall, the three-group salience hierarchy
(\vsink{} $>$ \lsink{} $>$ ordinary)
observed in LLaVA-1.5 (Fig.~2) is reproduced.

\begin{figure}[t]
\centering
\includegraphics[width=\linewidth]{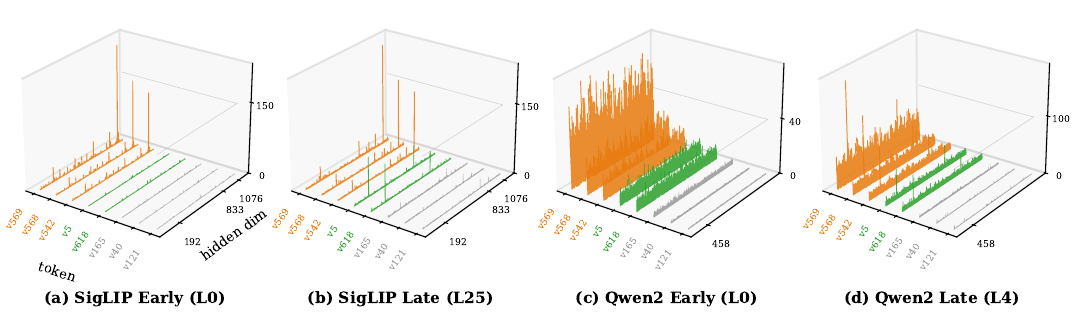}
\caption{\textbf{Activation patterns on LLaVA-OneVision-7B}
    (cf.\ Fig.~3).
    \textcolor{orange}{Orange}: V-sink,
    \textcolor[rgb]{0.17,0.63,0.17}{green}: L-sink,
    gray: ordinary.
    (a,\,b)~SigLIP first and last encoder layers.
    (c)~Qwen2 layer~0.
    (d)~Qwen2 layer~4.
    The corresponding spatial overlay is shown in Figure~\ref{fig:ov_overlay}.}
\label{fig:ov_3d}
\end{figure}

\begin{figure}[t]
    \centering
    \begin{minipage}[t]{0.46\linewidth}
        \centering
        \includegraphics[width=\linewidth]{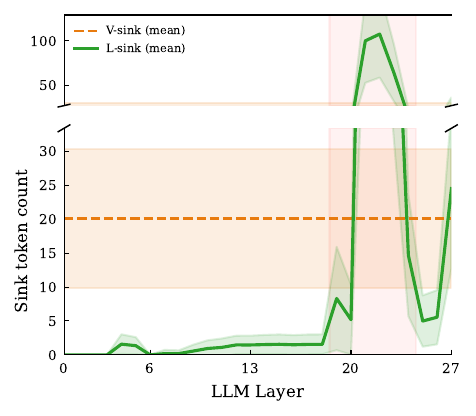}
        \captionof{figure}{\textbf{Sink count per layer
        on LLaVA-OneVision-7B}
        (cf.\ Figure~\ref{fig:sink_count}).
        Shaded band: L19--L24 ($y$-axis broken).}
        \label{fig:ov_sink_count}
    \end{minipage}\hfill
    \begin{minipage}[t]{0.50\linewidth}
        \centering
        \includegraphics[width=\linewidth]{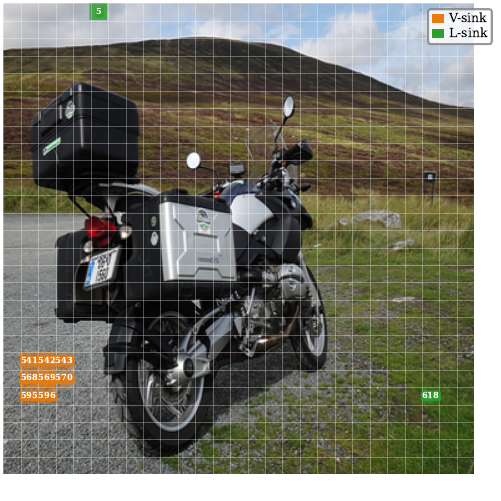}
        \captionof{figure}{\textbf{Spatial distribution of sink tokens
        on LLaVA-OneVision-7B}
        (cf.\ Figure~\ref{fig:image_overlay}).
        \textcolor{orange}{\textbf{Orange}}: V-sinks;
        \textcolor[rgb]{0.17,0.63,0.17}{\textbf{green}}: L-sinks.}
        \label{fig:ov_overlay}
    \end{minipage}
\end{figure}

\noindent\textbf{Sink count (Figure~\ref{fig:ov_sink_count}).}\quad
V-sinks (orange, dashed) remain constant at ${\sim}$20 across all layers.
L-sinks (green, solid) emerge at L4 and stay at ${\sim}$1--5 through L18.
The shaded band marks L19--L24,
where the count spikes due to residual norm growth
(the $y$-axis is broken to accommodate values reaching ${\sim}$100).
Counts recover at L25--L26,
and L27 shows a slight increase relative to L25--L26
but remains far below the L19--L24 peak.

\noindent\textbf{Activation patterns (Figure~\ref{fig:ov_3d}).}\quad
The 3D activation profiles reveal two architecture-specific findings.

\noindent\emph{(i) V-sink massive activations from the first encoder layer.}\quad
In CLIP-ViT-L/14,
massive activations at dim~650
emerge at an intermediate encoder layer
and grow through subsequent layers~\cite{sun2024massive}.
In SigLIP-SO400M, massive activations at dim~1076
are already present at the first encoder layer (panel~a)
and persist with similar magnitude through the last encoder layer (panel~b).

\noindent\emph{(ii) L-sinks with elevated ViT-side norms.}\quad
In LLaVA-1.5, L-sinks are indistinguishable from ordinary tokens
at the ViT output
and only develop sink characteristics inside the LLM
(Section~3.2 of the main paper).
In LLaVA-OneVision, the tokens that eventually become L-sinks
are already visually distinct at the SigLIP output:
panels~(a,\,b) show that these tokens (green)
carry elevated activations at dimensions other than dim~1076
(e.g., dims~192, 833),
which develop through the encoder depth
but do not reach $\tau_{\mathrm{vit}}$ at dim~1076,
so they are not classified as V-sinks.
The MLP projector amplifies these elevated norms:
at Qwen2 layer~0 (panel~c),
V-sinks (orange) enter with the largest norms,
L-sinks (green) enter with intermediate norms,
and ordinary tokens (gray) enter with the lowest.
By layer~4 (panel~d),
the FFN has written massive activations at dim~458
for both V-sinks and L-sinks,
completing L-sink formation through the same
early-layer mechanism as in LLaVA-1.5~\cite{sun2024massive, kang2025see}.

\medskip
\noindent\textbf{Summary.}\quad
The V-sink/L-sink/ordinary three-group partition
and its associated salience hierarchy
both generalize to LLaVA-OneVision-7B
despite substantial architectural differences.
The one notable distinction is that
L-sinks in LLaVA-OneVision already carry elevated norms
from secondary ViT dimensions,
whereas in LLaVA-1.5 they are norm-indistinguishable from ordinary tokens.
In both architectures, however,
L-sinks ultimately develop massive activations at the LLM's sink dimensions
and function as attention sinks within the three-group taxonomy.
The stable identification range (L4--L18)
suggests that layer-wise intervention,
as conducted for LLaVA-1.5 in Section~5,
would be feasible within this range.
We verify this via oracle key-gating in Section~\ref{sec:supp_cross_oracle}.
 
\subsection{Oracle Intervention on LLaVA-OneVision}
\label{sec:supp_cross_oracle}
 
The oracle grid search in Section~3.4 revealed that
the optimal gating direction and magnitude are
both task-dependent (Observation~1) and layer-dependent (Observation~2)
on LLaVA-1.5.
A natural question is whether these patterns are specific to
the CLIP-ViT-L/14~+ LLaMA-2 combination
or reflect a more general property of
layer-wise information routing in LVLMs.
To answer this, we replicate the oracle key-gating sweep
on LLaVA-OneVision-7B.
 
\noindent\textbf{Setup.}\quad
We apply the same intervention as Section~3.4:
at each target layer,
the key vectors of V-sink tokens are scaled by $\rho_{\mathrm{vit}}$
and those of remaining visual tokens by $1 - \rho_{\mathrm{vit}}$.
We sweep $\rho_{\mathrm{vit}} \in \{0.0, 0.1, \ldots, 1.0\}$
at seven layers sampled at approximately uniform intervals
(every 3--4 layers) across the 28-layer Qwen2 decoder:
L2, L5, L9, L13, L16, L19, and L23.
Each configuration is evaluated on
MMStar (6~subtasks) and CVBench (4~subtasks),
matching the LLaVA-1.5 evaluation protocol.
For reference, LLaVA-OneVision's unmodified baselines are
MMStar~61.9\% and CVBench~77.2\%
(vs.\ 33.3\% and 57.3\% for LLaVA-1.5;
per-subtask breakdown in Table~\ref{tab:ov_baseline}).
 
\begin{table}[t]
\centering
\caption{\textbf{Baseline accuracy~(\%) of LLaVA-1.5-7B and LLaVA-OneVision-7B.}}
\label{tab:ov_baseline}
\small
\setlength{\tabcolsep}{3pt}
\begin{tabular}{@{}l cccccc cccc@{}}
\toprule
& \multicolumn{6}{c}{MMStar} & \multicolumn{4}{c}{CVBench} \\
\cmidrule(lr){2-7} \cmidrule(lr){8-11}
& CP & FG & IR & LR & Math & ST & Cnt & Dep & Dist & Rel \\
\midrule
LLaVA-1.5 & 63.9 & 24.7 & 39.3 & 27.7 & 26.9 & 17.2 & 47.8 & 67.0 & 48.0 & 66.3 \\
LLaVA-OV  & 74.9 & 52.8 & 72.7 & 58.4 & 57.6 & 55.2 & 69.0 & 83.2 & 75.5 & 80.9 \\
\bottomrule
\end{tabular}
\end{table}
 
\begin{figure}[t]
\centering
\includegraphics[width=\linewidth]{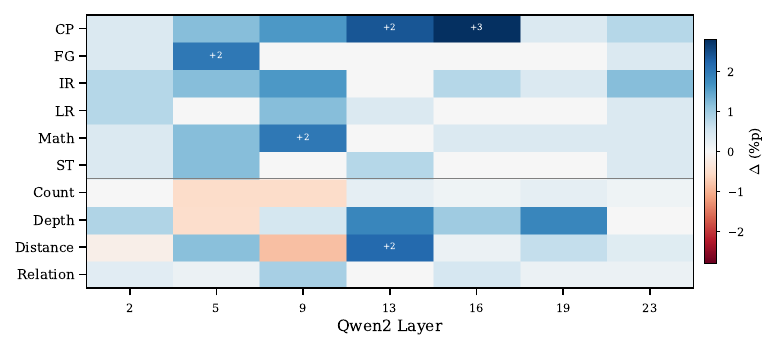}
\caption{\textbf{Oracle key-gating sweep on LLaVA-OneVision-7B}
    (cf.\ Figure~\ref{fig:oracle_vs_learned}).
    Each cell reports the best accuracy change~(\%p)
    among 11~swept gate values at that layer.
    Seven layers are sampled at uniform intervals
    across the 28-layer Qwen2 decoder.}
\label{fig:ov_oracle}
\end{figure}
 
\noindent\textbf{Results.}\quad
Figure~\ref{fig:ov_oracle} shows the per-subtask oracle results.
Despite LLaVA-OneVision's substantially stronger baseline
(Table~\ref{tab:ov_baseline}),
positive oracle gains exist at every sampled layer
(6--8 out of 10 subtasks per layer).
The heatmap reveals two patterns consistent with
those observed on LLaVA-1.5.
First, the effect is task-dependent (Observation~1):
even at the same layer,
the effect of the optimal gate value differs across subtasks,
with some subtasks improving while others remain flat or degrade.
Second, the effect is layer-dependent (Observation~2):
for a given subtask,
layers at which intervention yields a substantial gain
are distinct from those at which it has little effect.
These patterns indicate that
the layer-wise utilization of sink tokens
and its task- and layer-dependent nature,
identified on the CLIP-ViT-L/14~+ LLaMA-2 architecture,
are generally observable on the
SigLIP~+ Qwen2 combination as well.

\section{Extending LSG to Other Architectures}
\label{sec:cross_arch_port}

Appendix~\ref{sec:supp_cross} shows that the \vsink{}/\lsink{} \emph{taxonomy}
generalizes to LLaVA-OneVision. Here we ask whether \emph{LSG} carries over to other vision-encoder and
LLM backbone combinations. We study two LLaVA architecture variants:
Phi-3-mini~\cite{abdin2024phi3} with
CLIP-ViT-L/14\footnote{\url{https://huggingface.co/xtuner/llava-phi-3-mini-hf}}
and Qwen2.5-3B with
SigLIP\footnote{\url{https://huggingface.co/Zhang199/TinyLLaVA-Qwen2.5-3B-SigLIP}}~\cite{zhou2024tinyllava}.  

\noindent\textbf{Setup.}\quad
For each architecture we identify \vsinks{} with the same activation-based
criterion as the main paper (Eq.~\ref{eq:sink_def}); the per-architecture sink
dimension and threshold are selected by the same procedure as
Appendices~\ref{sec:supp_sink_id} and~\ref{sec:supp_cross_setup}. All metrics
use the same protocol as the main paper. We report 
an \emph{oracle} per-layer sink-gate sweep and a \emph{learned} single-layer LSG.  

\begin{table}[t]
\centering
\caption{Cross-architecture downstream performance. $\Delta$ (\%p) over each
architecture's own baseline (Phi-3+CLIP: 44.84/54.10; Qwen2.5+SigLIP:
44.04/59.68). Or = oracle sweep at the layer; LSG = a learned single-layer
2-group gate.}
\label{tab:cross_arch_lsg}
\small
\setlength{\tabcolsep}{5pt}
\begin{tabular}{lcccc}
\toprule
 & \multicolumn{2}{c}{Phi-3 (3.8B)+CLIP} & \multicolumn{2}{c}{Qwen2.5 (3B)+SigLIP} \\
\cmidrule(lr){2-3} \cmidrule(lr){4-5}
 & Or@L20 & LSG@L20 & Or@L4 & LSG@L4 \\
\midrule
$\Delta$\,MMStar  & $+0.73$ & $+0.21$ & $+1.67$ & $+0.42$ \\
$\Delta$\,CVBench & $+1.95$ & $+0.36$ & $+1.10$ & $+0.89$ \\
\bottomrule
\end{tabular}
\end{table}

\begin{figure}[t]
\centering
\begin{subfigure}{\linewidth}
  \centering
  \includegraphics[width=\linewidth]{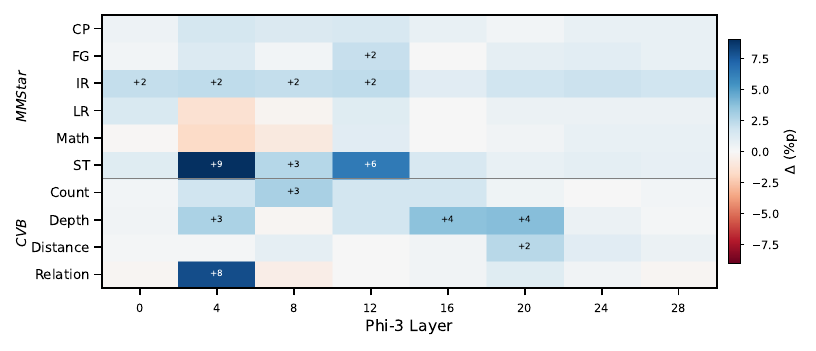}
  \caption{Phi-3+CLIP}
  \label{fig:cross_arch_oracle_phi3}
\end{subfigure}

\vspace{4pt}

\begin{subfigure}{\linewidth}
  \centering
  \includegraphics[width=\linewidth]{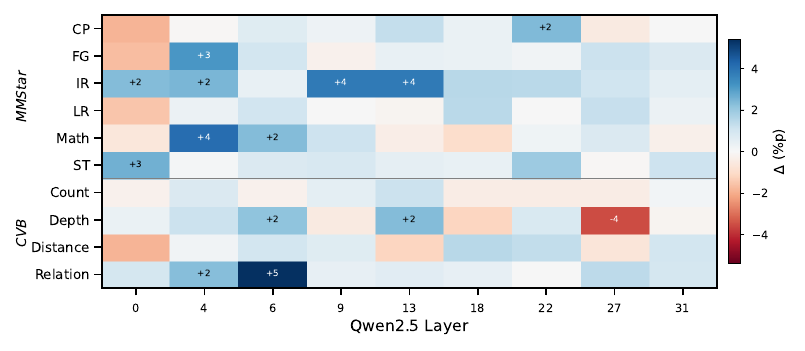}
  \caption{Qwen2.5+SigLIP}
  \label{fig:cross_arch_oracle_qwen}
\end{subfigure}
\caption{\textbf{Oracle key-gating sweep on two further architectures.}
Each cell reports the best accuracy change (\%p) among 11 swept gate values at
that layer, per sub-task.}
\label{fig:cross_arch_oracle}
\end{figure}

\noindent\textbf{Findings.}\quad
On both architectures the oracle sweep reproduces the main paper's two
observations: the effect of \vsink{} gating is task-dependent (Obs.~1) and
layer-dependent (Obs.~2), shown per sub-task in
Figures~\ref{fig:cross_arch_oracle_phi3} and~\ref{fig:cross_arch_oracle_qwen}.
A learned single-layer LSG produces same-sign, positive $\Delta$ on both
benchmarks at the matched layer: on Phi-3+CLIP, $+0.21$ on MMStar and $+0.36$
on CVBench at L20; on Qwen2.5+SigLIP, $+0.42$ and $+0.89$ at L4. These are
reported alongside the oracle upper bound in Table~\ref{tab:cross_arch_lsg}. We take these as preliminary evidence that LSG ports across architectures.

\end{document}